\documentclass{ieeeaccess}
\usepackage{cite}
\usepackage{amsmath,amssymb,amsfonts}
\usepackage{algorithmic}
\usepackage{graphicx}
\usepackage{textcomp}
\usepackage{color,soul}
\usepackage{booktabs}
\usepackage[export]{adjustbox}
\usepackage{multirow}
\usepackage{url}
\usepackage{hyperref}  
\usepackage[utf8]{inputenc}
\DeclareMathOperator*{\argmax}{\arg\!\max} 
\DeclareMathOperator*{\argmin}{\arg\!\min} 
\usepackage[font=small,justification=justified]{caption}
\DeclareCaptionFont{ieeeblue}{\color{accessblue}}
\makeatletter
\newcommand{\ssymbol}[1]{^{\@fnsymbol{#1}}}
\makeatother
\def\BibTeX{{\rm B\kern-.05em{\sc i\kern-.025em b}\kern-.08em
    T\kern-.1667em\lower.7ex\hbox{E}\kern-.125emX}}

\renewcommand\hl[1]{#1}
\begin{document}
\history{}
\doi{10.1109/ACCESS.2020.3007337}

\title{\Large{Radial Basis Function Networks for Convolutional Neural Networks to Learn Similarity Distance Metric and Improve Interpretability}}
\author{Mohammadreza Amirian\authorrefmark{1, 2}, and
Friedhelm Schwenker\authorrefmark{2}}
\address[1]{Institute of Applied Information Technology, Zurich University of Applied Sciences, 8400 Winterthur, Switzerland (e-mail: \href{mailto:amir@zhaw.ch}{amir@zhaw.ch})}
\address[2]{Institute of Neural Information processing, Ulm University, 89081 Ulm, Germany (e-mail: \href{mailto:friedhelm.schwenker@uni-ulm.de}{friedhelm.schwenker@uni-ulm.de})}

\markboth
{Amirian \& Schwenker: RBFs for CNNs and Similarity Metric Learning}
{Amirian \& Schwenker: RBFs for CNNs and Similarity Metric Learning}

\corresp{Corresponding author: Mohammadreza Amirian (e-mail: \href{mailto:amir@zhaw.ch}{amir@zhaw.ch}).}

\begin{abstract}
Radial basis \hl{function} neural networks (RBFs) are prime candidates for pattern classification and regression and have been used extensively in classical machine learning applications. However, RBFs have not been integrated into \hl{contemporary} deep learning research and computer vision \hl{using conventional convolutional neural networks (CNNs)} due to their lack of adaptability with modern architectures. In this paper, we adapt \hl{RBF networks as a classifier on top of CNNs} by modifying the training process and introducing a new activation function to train modern vision architectures end-to-end for image classification. The specific architecture of RBFs \hl{enables the learning of} a similarity distance metric to compare and find similar and dissimilar images. Furthermore, we demonstrate that using an RBF classifier on top of any CNN architecture provides new human-interpretable insights about the decision-making process of the models. \hl{Finally, we successfully apply RBFs to a range of CNN architectures and evaluate the results} on benchmark computer vision datasets.
\end{abstract}


\titlepgskip=-15pt

\maketitle
\renewcommand{\figurename}{Figure}

\section{Introduction}

\begin{figure*}[htb]
    \centering
    \begin{minipage}{\textwidth}
    \includegraphics[width=\linewidth]{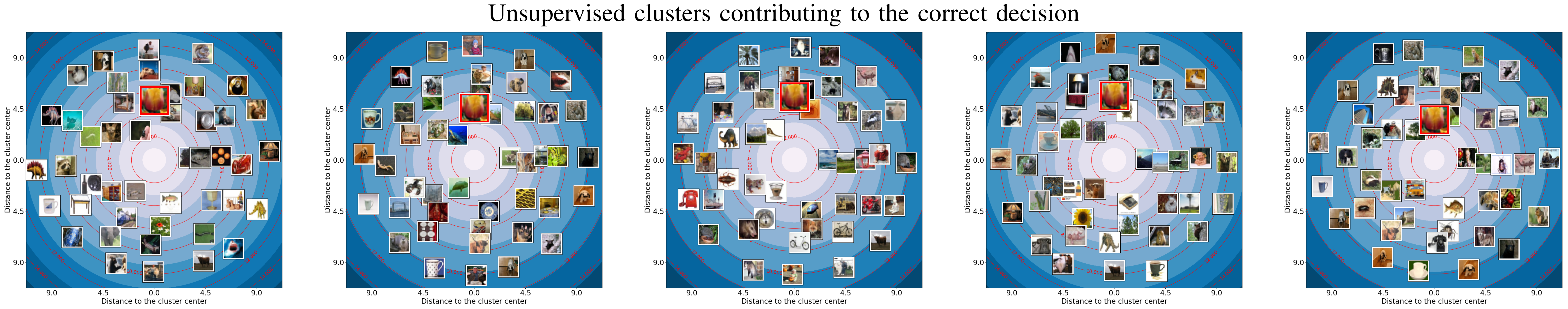} \\
    \includegraphics[width=0.5\linewidth]{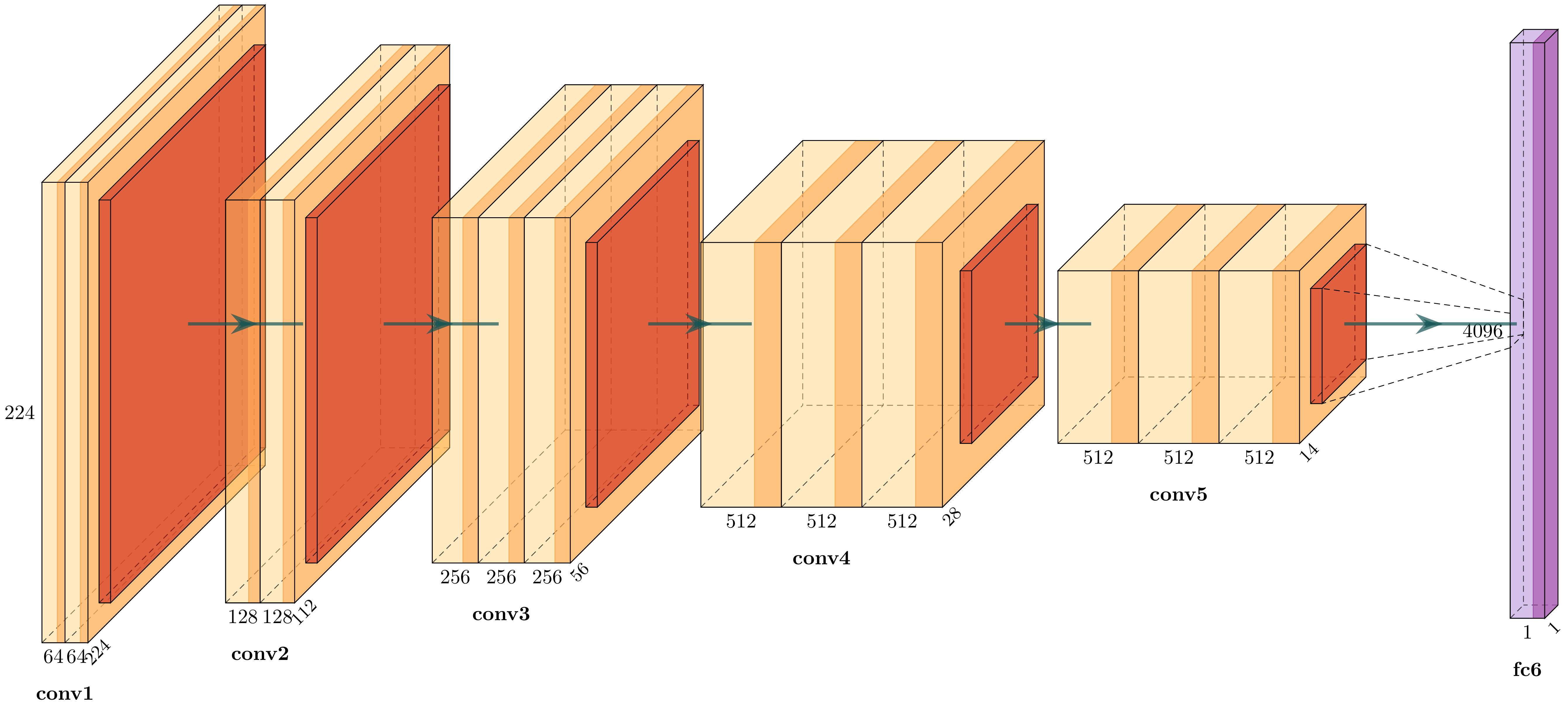}
    \includegraphics[width=0.10\linewidth]{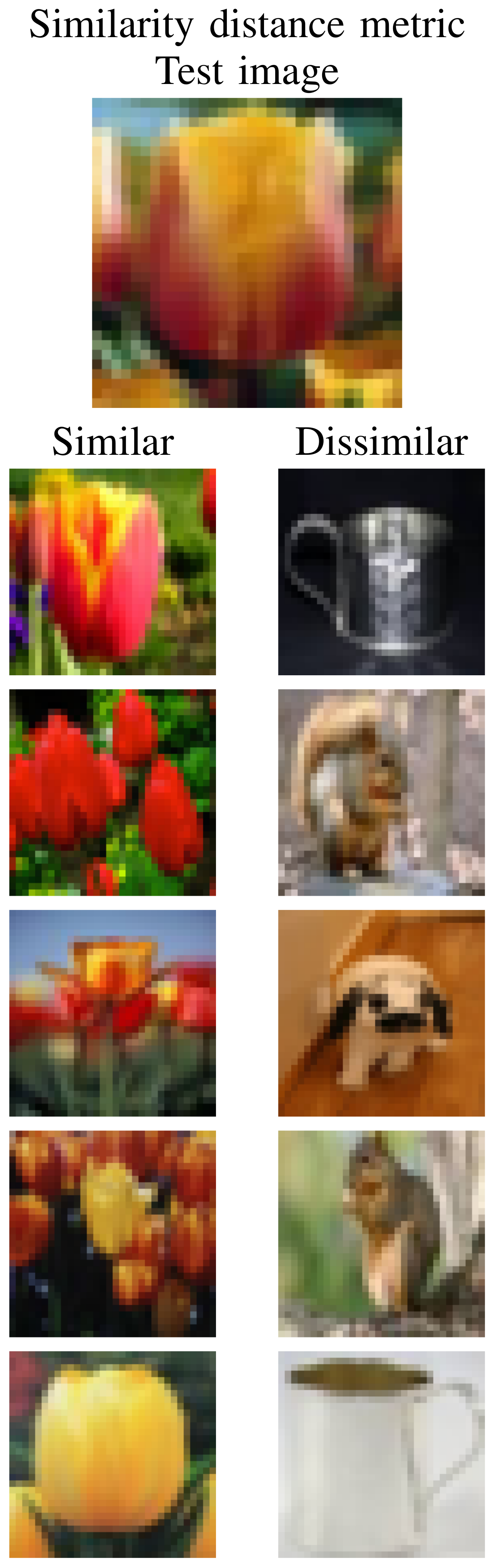}
    \includegraphics[width=0.3\linewidth]{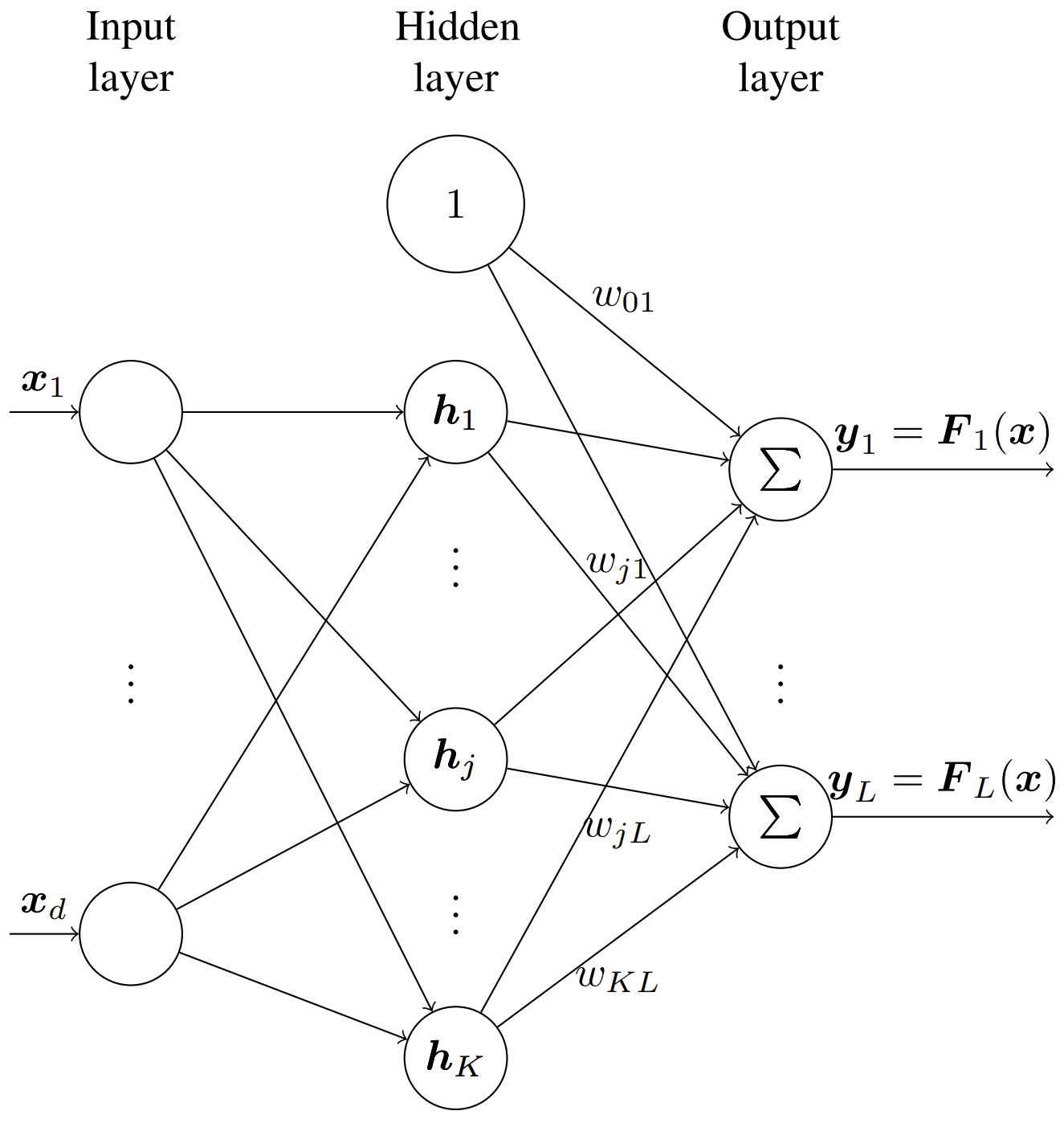} \\
    \includegraphics[width=\linewidth]{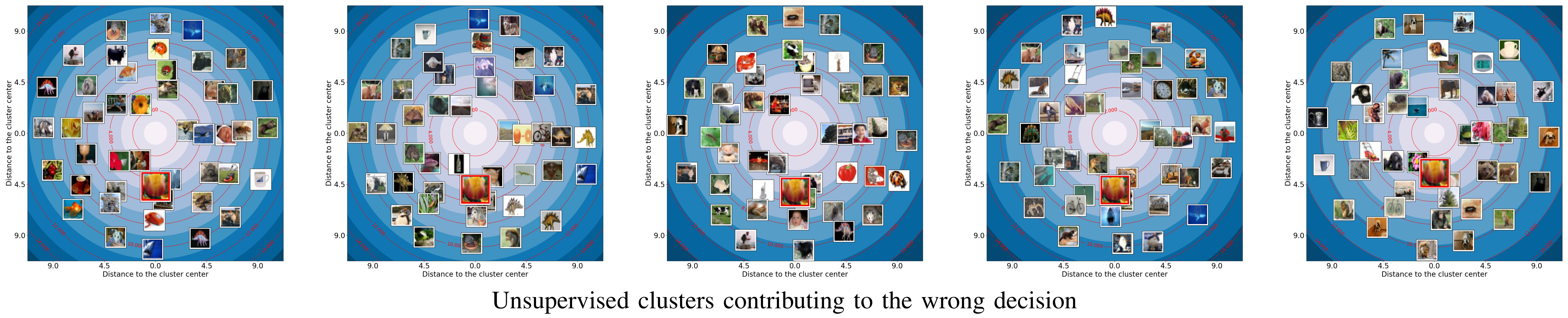} \\
    \end{minipage}
    \caption{A visual summary of the paper: Figures on the top and bottom rows visualize the position of a test image in the unsupervised clusters. \hl{The output of a CNN backbone is connected to the input of an RBF through a fully connected layer. The input features of the RBFs are referred to as embeddings in this research work. The embeddings of each image are compared to cluster centers with a trainable similarity distance metric. The same distance metric can be used to find images similar to a test sample amongst training images (visualized in the table in the middle row). The RBFs apply an activation function to the distance of the training images from the cluster centers to compute activation values. The output layer of the RBF is optimized for classification based on these activation values. We train the entire CNN-RBF architecture end-to-end.}}
    \label{fig:CNN-RBF}
\end{figure*}

Inspired by the locally tuned response of biological neurons, Broomhead and Lowe introduced radial basis \hl{function} neural networks (RBFs) in 1988 \cite{broomhead1988multivariable}. The modeling concept behind RBFs is a combination of unsupervised and supervised learning for pattern classification and regression. However, RBFs have not been integrated into the \hl{contemporary approaches to computer vision using Convolutional Neural Networks (CNNs) so far due to} structural deficiencies. This paper presents developments in a new area of research and lays the foundation for using RBFs in deep learning and computer vision by modifying their architecture and learning process. The results demonstrate that integrating RBFs into \hl{CNN models for computer vision} provides both a similarity distance metric \hl{as well as} an interpretable decision-making process.  

\hl{This research is motivated by the unique opportunities the RBF architectures introduce when used with CNN models. A new training process introduced for RBFs in this paper provides the opportunity of using labeled and unlabeled data by optimizing two loss functions combining supervised and unsupervised learning. The training process of RBF architectures employs a distance metric optimization that we propose to use as a similarity distance metric to find similar and dissimilar images. Additionally, this research proposes visualization techniques to illustrate the clusters and activations with training and test images to gain more insight into the reasoning behind the decisions made by the networks, thus improving interpretability. The contributions of this paper to computer vision literature can be summarized as follows:}        
\begin{itemize}
    \item \hl{Combining supervised and unsupervised learning.}
    \item Learning a similarity distance metric to find similar images.
    \item Improving the interpretability of decision-making.
\end{itemize}

\hl{Despite the advantages of combining RBFs to modern CNN architectures, there are} two factors in the architecture and training process of RBFs hindering their integration into CNNs. First, the nonlinear activations and computational graphs of RBFs used in the literature prevent efficient gradient flow. Secondly, RBFs assume that the training features are fixed and the cluster centers are initialized accordingly. Nonetheless, CNN architecture learns the embeddings, which are used as features of RBFs. This paper tackles the limitations of the original RBFs and presents the following contributions to RBF literature:

\begin{itemize}
    \item Introducing a quadratic activation function and a linear computational graph for end-to-end learning.
    \item Adding an unsupervised loss term to update the cluster centers in the training process with the learned embeddings.
    \item \hl{Applying the RBFs to computer vision in a first attempt using deep CNN architectures.} 
\end{itemize}


The remainder of the paper covers the related works in Section \ref{sec:related} followed by the theoretical background of RBFs in Section \ref{sec:RBF}.
We then present our original research and contributions in Section \ref{sec:CNN-RBF} with our proposed modifications to RBFs followed by a visual explanation of the new proposed training and decision-making process in Section \ref{sec:training}.
The experimental results of applying the proposed RBF-CNN architectures using a range of CNN backbones on benchmark datasets are presented in Section \ref{sec:experimental}.
The potential contributions of our proposed similarity distance metric on the field of computer vision to enhance the transparency of the decision making process is demonstrated in Section \ref{sec:interpret}.
We finally present our conclusions in \ref{sec:conclusion}.

\section{Related Works}
\label{sec:related}

Research into optimizing RBF architectures has followed two \hl{approaches}. The first \hl{appraoch} concentrates on the training process and initialization of the networks while the second \hl{approach aims to find better optimized} activation functions. This paper presents improvements to both \hl{approaches and integrates the RBFs into contemporary} vision models using CNNs. 

RBFs were originally introduced as a supervised approach for classification and regression \hl{tasks}. Broomhead and Lowe proposed to draw the cluster centers either from a uniform distribution or randomly from the training samples and then \hl{optimizing} the output weights using a pseudo-inverse analytic solution~\cite{broomhead1988multivariable}. Initializing the cluster centers randomly and only training the output weights is a one-phase training process for RBFs. Two-phase training for RBFs uses various methods to initialize the cluster centers as well as optimizing the output weights. \hl{Research since 1988 has used supervised and unsupervised methods to initialize the centers}. Moody and Darken proposed an unsupervised algorithm to initialize \hl{these cluster centers}~\cite{moody1989fast} while Schwenker et al. proposed supervised vector quantization~\cite{schwenker1994similarities}. \hl{Decision} trees were used to find centers independently by~\cite{kubat1998decision} and~\cite{schwenker2000initialisation} before training the output weights. Finally, Schwenker et al. proposed a third phase to optimize the entire RBF network end-to-end including output weights, cluster center, and parameters of activation functions using gradient descent~\cite{schwenker2001three}.   

All of these methods for cluster center initialization assume a fixed feature space for the input layer. However, CNNs learn the embeddings automatically and develop the feature space of the images during the training process. Therefore, we suggest optimizing an unsupervised learning loss during the training to cope with \hl{this change in} the feature space. This work differs from previous research as it combines supervised and unsupervised learning by optimizing two \hl{separate} losses using gradient descent.

Various applications and implementations have motivated several activation functions presented in the literature of RBFs~\cite{du2002fast}. The Gaussian function is the kernel encouraged by modeling the data through a multivariate Gaussian distribution~\cite{broomhead1988multivariable}. Other functions adopted in the RBF architecture include \hl{linear kernels, thin-plate splines, logistic functions, and multiquadratic functions}~\cite{franke1979critical, poggio1990networks,liao2003relaxed,chen1990practical}.
 Hardy's multiquadratic functions motivate an activation function for RBFs used by Karimi et al. and Zhao et al.~\cite{karimi2020generalized, zhao2019prediction}. Du et al. proposed a kernel for digital signal processing (DSP) units \ref{eq:kernels7}. In this paper, we suggest a quadratic kernel to build a linear computational graph for efficient gradient flow and \hl{integrate RBFs} for end-to-end training with CNN architectures.

Besides the mature fundamental research, RBFs have been applied to a broad range of applications for pattern classification and regression in recent years. Nicodemou et al. \hl{used} RBF networks for 3D hand pose estimation~\cite{nicodemou2020single}, Dehghan and Mohammadi \hl{estimated} a numerical solution for Fokker-Planck differential equations with RBFs~\cite{dehghan2014numerical}, Li et al. used sparse multiscale RBFs for seizure detection in EEG signals~\cite{li2019epileptic}, and Zhao et al. predicted interfacial interactions by training RBFs~\cite{zhao2019prediction}. Furthermore, Geng et al. introduced deep RBF networks and applied the method to food safety inspection data hl{and} RBFs are used to train models for classification and regression in discrete and continuous pain quantification~\cite{amirian2016using}.

\hl{RBFs have been used for computer vision tasks and image classification as well. Friedhelm et al. used raw images as feature vectors to classify hand-written digits}~\cite{schwenker2001three}. \hl{Er et al. extracted the features from facial images using principle component analysis (PCA) and processed these features using Fisher's linear discriminant (FLD) technique before classifying the patterns using RBFs}~\cite{er2002face}. \hl{However, the successful rise of the modern conventional neural networks, such as LeNet-5}~\cite{lecun2015lenet} \hl{and AlexNet}~\cite{russakovsky2015imagenet}\hl{, led to a paradigm shift from using hand-crafted features to automated deep convolutional feature and representation learning using CNNs. In recent years, most computer vision tasks, like facial recognition}~\cite{masi8614364survey}, \hl{are dominated by modern CNN architectures as they present superior performance compared to classical methods for image processing. To the best of our knowledge, this paper presents the first attempt to integrate RBFs into modern CNN architectures for computer vision.}

\hl{This research work relates to literature focusing on deep metric learning since RBFs optimize a similarity distance metric automatically during training based on their architecture. Euclidean distance, Mahalanobis distance and cosine similarity have been used to evaluate similarity between the embeddings (extracted features from CNNs) of two images in the literature}~\cite{hoffer2015deep, xing2003distance, wojke2018deep}\hl{. Researchers have applied different strategies and loss functions to optimize these similarity metrics for same-class images while also maximizing the distance of different-class images. The research in this area concentrates on the training process and the design of a loss function which brings similar images closer in the embedding space based on a similarity measure. Hu et al. proposed to minimize the inter-class scores and maximize the intra-class scores based on Euclidian distances}~\cite{hu2015deep}\hl{. Hoffer and Ailon suggested optimizing a similarity-based loss function defined for selected triplets of images}~\cite{hoffer2015deep}\hl{. Song et al. used the pairwise distances between images of an entire batch and proposed a structured loss function for metric learning}~\cite{oh2016deep}. \hl{Similar research work aimed at optimizing angular distance, cosine distance and large-margin Euclidean distance of similar and dissimilar samples}~\cite{wang2017deep, wojke2018deep, cheng2018deep}. 

\hl{This paper presents a method to retrieve a ranked list of similar and dissimilar images, which leads to visually appealing results for similarity metric learning. However, the proposed similarity metric learned by the RBFs does not require any complicated triplet sample section or loss design. The presented results are obtained using a typical supervised loss function for classification (softmax cross-entropy). Furthermore, RBFs can not only optimize for Euclidean and Mahalanobis distances, but also for the entire covariance matrix.}     
\section{Radial Basis Function Networks}

\begin{figure*}{}
    \centering
    \includegraphics[width=0.4\textwidth, valign=t]{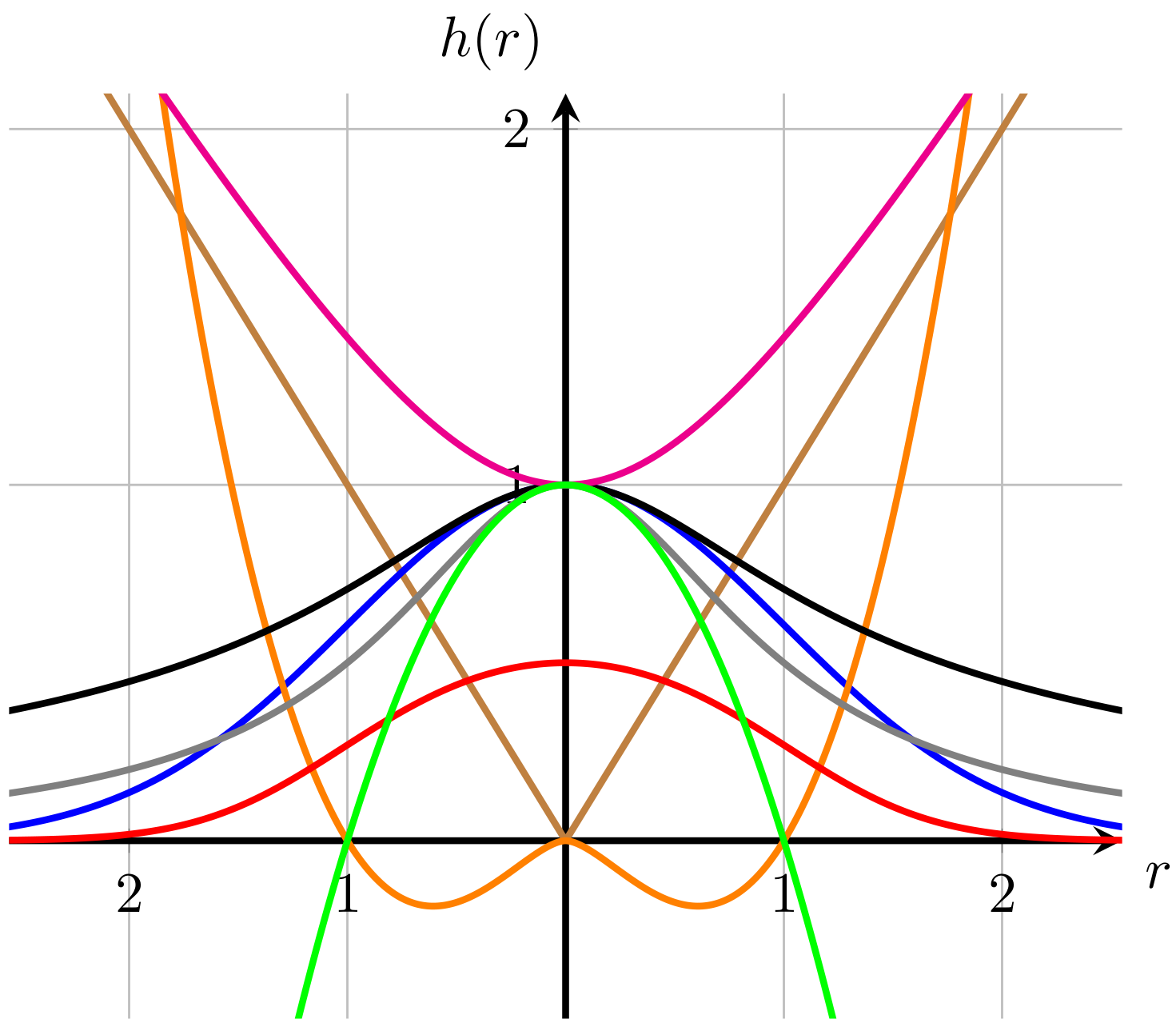}
    \hspace{-0.35cm}
    \raisebox{-0.75cm}{\includegraphics[width=0.17\textwidth, valign=t]{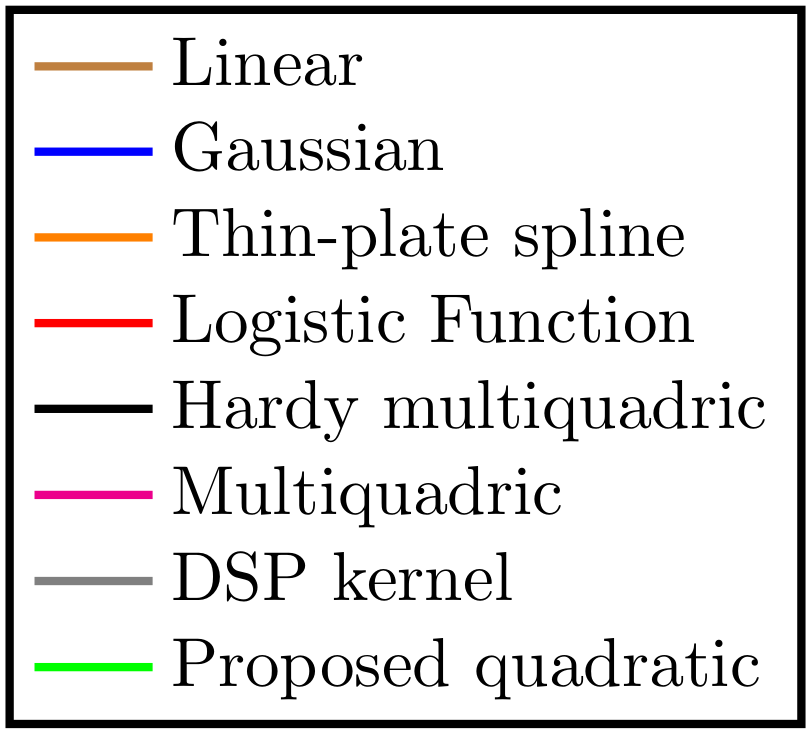}}
    \includegraphics[width=0.4\textwidth, valign=t]{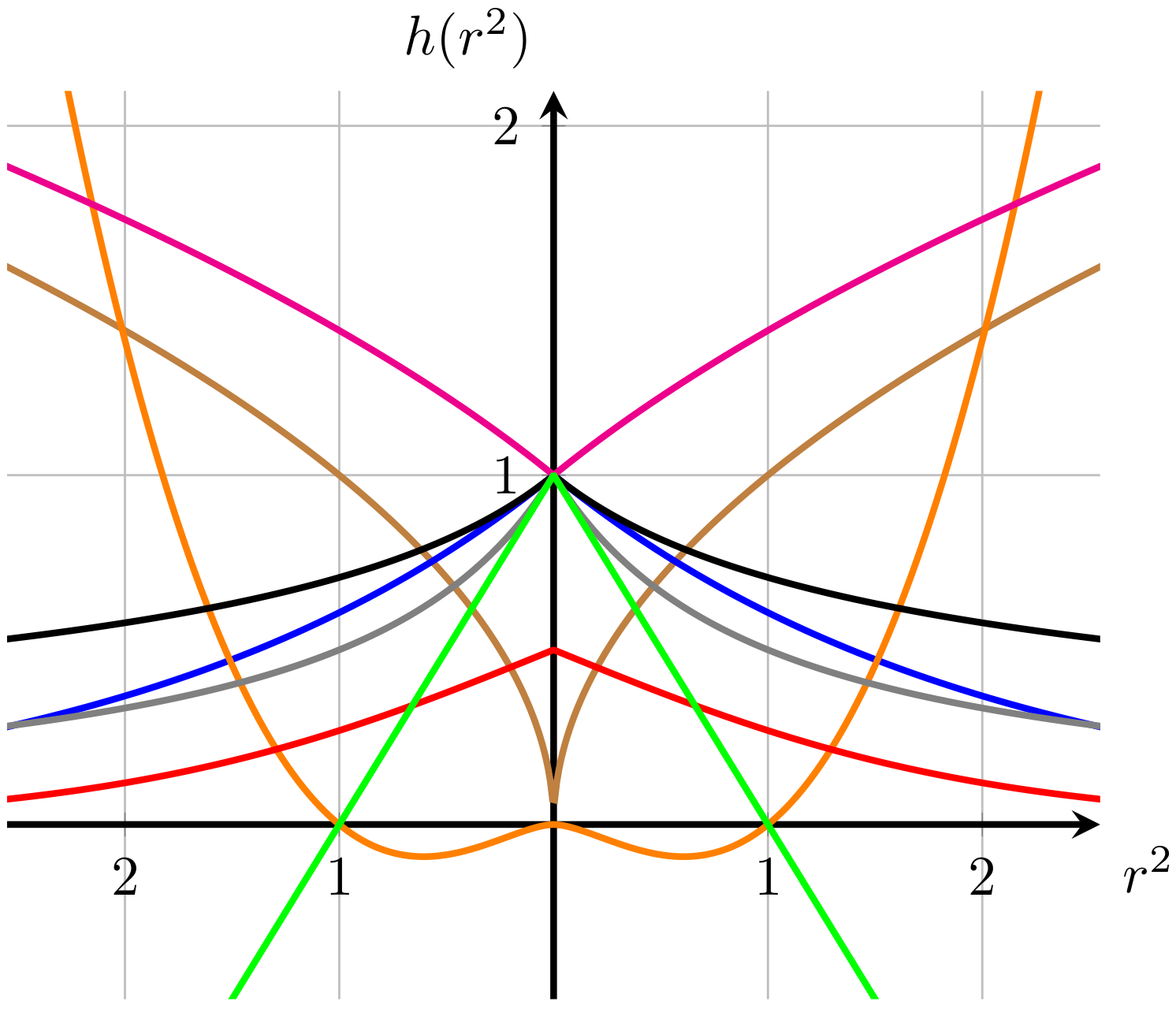}
    \caption{Activation functions for RBF networks. We used the following parameters for all of the kernels: $\sigma=1$, $\alpha=1/2$, and $\beta=1/2$. The proposed quadratic activation kernel is linear based on the $r^2$. Consequently, the CNN goes through a completely linear forward path and thus computes the gradient during backpropagation efficiently.}
 	\label{fig:activation}
\end{figure*}

\label{sec:RBF}
In this section, we briefly review and explain the theoretical foundation of radial basis function networks. RBFs are presented in the literature as a global approximation method for learning a mapping $\boldsymbol{F}$ from a given feature space with \hl{dimensionality} $d$ to label a space with \hl{dimensionality} $K$ ($\boldsymbol{F}: {\rm I\!R}^d \rightarrow {\rm I\!R}^K$) \cite{broomhead1988multivariable}. In this paper, the \hl{function} $\boldsymbol{F}$ of features $\boldsymbol{x}$ approximates the one-hot encoded labels $\boldsymbol{y}$. The features used in this work to train the RBFs are the embeddings of deep (CNNs) to predict the class labels using end-to-end optimization. However, we use a fully connected layer between CNN architectures and RBFs to provide compatibility between the two architectures and prevent overfitting. The architecture of the RBF consists of an input layer, a single trainable hidden layer with $C$ cluster centers ($\boldsymbol{c}_j$) \ref{fig:CNN-RBF}, and an output layer. 

During the evaluation, also termed inference in the \hl{deep learning} architecture, the RBF computes a distance between embeddings of deep CNNs and the cluster centers and applies an activation function to \hl{this} distance. \hl{The network outputs are then} computed by multiplying the weights of the output layer \hl{with} the activation values. This evaluation process \hl{formally defined as}:
\begin{flalign}\label{eq:distance}
r^2 = 
(\boldsymbol{x}-\boldsymbol{c}_j)^T\boldsymbol{R}_j(\boldsymbol{x}-\boldsymbol{c}_j) &&
\end{flalign}
\begin{flalign}\label{eq:output}
\boldsymbol{y}_k = \boldsymbol{F}_k(\boldsymbol{x}) = \sum_{j=1}^{C}w_{jk} h(\parallel \boldsymbol{x}^\mu-\boldsymbol{c}_j \parallel^2_{\boldsymbol{R}_j}) + w_{0k} &&
\end{flalign} 
where $r$ represents the distance, $R_j$ is the positive definite covariance matrix \hl{(trainable distance)}, $T$ denotes the matrix transposition, $ w_{jk} $ shows the weights of the output layers, $h$ is the activation function, and $w_{0k}$ are the biases. In these Equations, $\mu$, $j$, and $k$ enumerate the number of samples, cluster centers, and classes, respectively. Trainable parameters in Equation \ref{eq:distance} and \ref{eq:output} are the output weights, cluster centers, and covariance matrix. 

Optimizing the RBF networks with an identity covariance matrix results in training in Euclidean space. It is possible to optimize a Mahalanobis distance \cite{de2000mahalanobis} by training the main diagonal on the covariance matrix. Any arbitrary distance metric can be trained by optimizing the entire covariance matrix while projecting the matrix to the space of positive definite matrices. The distance $r$ computed in Equation \ref{eq:distance} is not only a measure of the proximity of an image to a cluster center but can also be used to compare images and find similar and dissimilar images in the embeddings space.

The linear and nonlinear activation functions used in RBFs are as follows \cite{poggio1990networks,liao2003relaxed,chen1990practical}: 

\begin{flalign}
	&\text{Linear}:            & h(r) &= r                                           && \label{eq:kernels1}\\
	&\text{Gaussian}:          & h(r) &= e^{-r^2/2\sigma^2}                          && \label{eq:kernels2}\\
	&\text{Thin-plate spline}: & h(r) &= r^2 \ln{r}                                  && \label{eq:kernels3}\\
	&\text{Logistic function}: & h(r) &= \frac{1}{1+e^{(r^2-r^2_0)/\sigma^2}}        && \label{eq:kernels4}\\
	&                          & h(r) &= \frac{1}{(r^2+\sigma^2)^\alpha},~\alpha > 0 && \label{eq:kernels5}\\
	&                          & h(r) &= (r^2+\sigma^2)^\beta,~0 < \beta < 1         && \label{eq:kernels6}\\ 
	&                          & h(r) &= \frac{1}{1+r^2/\sigma^2}                    && \label{eq:kernels7}
\end{flalign}

In addition to the standard machine learning activation kernels in Equations \ref{eq:kernels1} to \ref{eq:kernels4}, the kernel presented in Equation \ref{eq:kernels5} is derived \hl{from the} generalized Hardy's multiquadratic function \cite{franke1979critical} and results in the same function for $\alpha=1/2$. Du et al. \cite{du2002fast} proposed the kernel in Equation \ref{eq:kernels7} because of its convenience for \hl{implementation} on DSP units. Various activation functions for RBF are depicted in Figure \ref{fig:activation}.

The complete process of training RBFs was introduced by Schwenker et al. \cite{schwenker2001three} \hl{as three phase process}:

\textbf{Unsupervised learning}: This step is aimed at finding cluster centers that are representative of the data. The k-means \cite{anderberg1973cluster} clustering algorithm is widely used for this purpose. k-means iteratively finds a set of cluster centers and minimizes the overall distance between cluster centers and members over the entire dataset. The target of the k-means algorithm can be written in the following form:
\begin{flalign}\label{eq:clus}
    \text{Loss}_{unsupervised} = \sum_{j=1}^{K}\sum_{\boldsymbol{x}^\mu\in\vartheta_j}^{} \parallel \boldsymbol{x}^\mu-\boldsymbol{c}_j \parallel^2 &&
\end{flalign} 
where $\boldsymbol{x}^\mu\in\vartheta_j$ denotes the members of the $j^{th}$ cluster shown by $\vartheta_j$. 

\textbf{Computing weights}: The output weights of an RBF network can be computed using a closed-form solution. The matrix of activation of the samples is defined from the training set ($H$) as follows:
\begin{flalign}\label{eq:activationMatrix}
    \boldsymbol{H} = h(\parallel \boldsymbol{x}^\mu-\boldsymbol{c}_j \parallel^2_{\boldsymbol{R}_j})_{\mu=1,~...~,M, j=1,~...~,C} &&
\end{flalign}

Based on Equation \ref{eq:output}, the matrix of output weights ($W$), \hl{which} estimates the matrix of labels ($Y$), is computed by the following equation:
\begin{flalign}\label{eq:computeWeight}
    \boldsymbol{Y} \approx \boldsymbol{H}\boldsymbol{W} \Rightarrow \boldsymbol{W} \approx \boldsymbol{H}\ssymbol{2}\boldsymbol{Y} &&
\end{flalign}
where $\boldsymbol{H}\ssymbol{2}$ is the Moore–Penrose pseudo-inverse matrix \cite{penrose1955generalized} of $\boldsymbol{H}$ and is computed as:
\begin{flalign}\label{eq:inverse}
    \boldsymbol{H}\ssymbol{2} = \lim\limits_{\alpha \to 0^+}(\boldsymbol{H}^T\boldsymbol{H} + \alpha \boldsymbol{I})^{-1} \boldsymbol{H}^T &&
\end{flalign}

\textbf{End-to-end optimization}: After initializing the RBF weights and cluster centers with a clustering algorithms such as k-means, it is possible to optimize the network end-to-end via backpropagation and gradient descent. Schwenker et al. computed the gradients of the loss function for a Gaussian activation function in \cite{schwenker2001three}.

\section{Adapting RBFs for CNNs}
\label{sec:CNN-RBF}

In this section, we propose using RBF classifiers on top of CNNs as depicted in Figure \ref{fig:CNN-RBF}. The deep embeddings are computed using standard convolutional layers and inception blocks. The deep embedding of the CNNs are flattened and fed to an RBF after using a fully connected layer in the architecture. The network ends with an output layer with softmax activation and is optimized end-to-end. Integrating the RBFs into deep structures and using them in conjunction with CNNs presents three challenges:

\textbf{Initialization}: Training the RBFs from scratch with randomly initialized weights using gradient descent is quite inefficient due to inappropriate initial cluster centers. The large initial distances in high dimensional spaces leads to small activation \hl{values and} the gradients attenuate considerably after the \hl{RBF hidden layer} during backpropagation. Therefore, we use the k-means algorithm to initialize the cluster centers before starting the training. Furthermore, computing the weights from Equation \ref{eq:computeWeight} is not feasible at the scale of computer vision problems such as ImageNet \cite{deng2009imagenet}, which has over $14$ million images and $1000$ classes. Hence, we randomly initialize the output layer and optimize it using gradient descent.

\textbf{Dynamic input features}: The input features of classical RBFs are fixed, but this assumption is not valid \hl{with respect to} CNNs. \hl{As} the embeddings of CNNs develop during the training process, the cluster centers initialized by the k-means algorithm are no longer optimal after a few epochs of training. We propose to optimize the k-means \hl{algorithm target with the} unsupervised loss \hl{defined} in Equation \ref{eq:clus} during the training process.

\textbf{Activation}: The nonlinear computational graph drawn by computing the distance in Equation \ref{eq:distance} and applying the activations in equations \ref{eq:kernels1}-\ref{eq:kernels7} leads to inefficient gradient flow. We target this challenge by introducing a new activation function.

\subsection{Introducing unsupervised learning loss}
\hl{In this section, we explain two modifications to classical RBFs to make them suitable for deep CNNs. First, we introduce an additional loss term to the RBF hidden layer. This term is based on the target function of the k-means algorithm defined in} Equation \ref{eq:clus} \hl{and continues in the unsupervised learning process during the development of embeddings. Secondly, we introduce a new quadratic kernel to build a linear computational graph for efficient optimization using backpropagation.}

The embeddings of CNNs change during the training process \hl{and} necessitates updating the cluster centers with an unsupervised loss. We introduce an additional term to the networks loss function based on the unsupervised target in Section \ref{sec:RBF} to optimize the cluster centers during training using the k-means loss in Equation \ref{eq:clus}.
 
The final loss of a CNN with RBFs as \hl{the} classifier (CNN-RBFs) is computed as
\begin{flalign} \label{eq:LossRBF}
    \text{Loss}_\text{rbf} = \text{Loss}_\text{supervised}+\lambda\times\text{Loss}_\text{unsupervised}  &&
\end{flalign}
where the classification loss $\text{Loss}_\text{classification}$ is any arbitrary loss function, for instance categorical cross entropy.

\renewcommand{\tablename}{Figure}
\setcounter{table}{2}
\begin{table*}[!ht]
     \begin{center}
     \begin{tabular}{c c c c c}
     \toprule
     
     \includegraphics[width=0.175\linewidth]{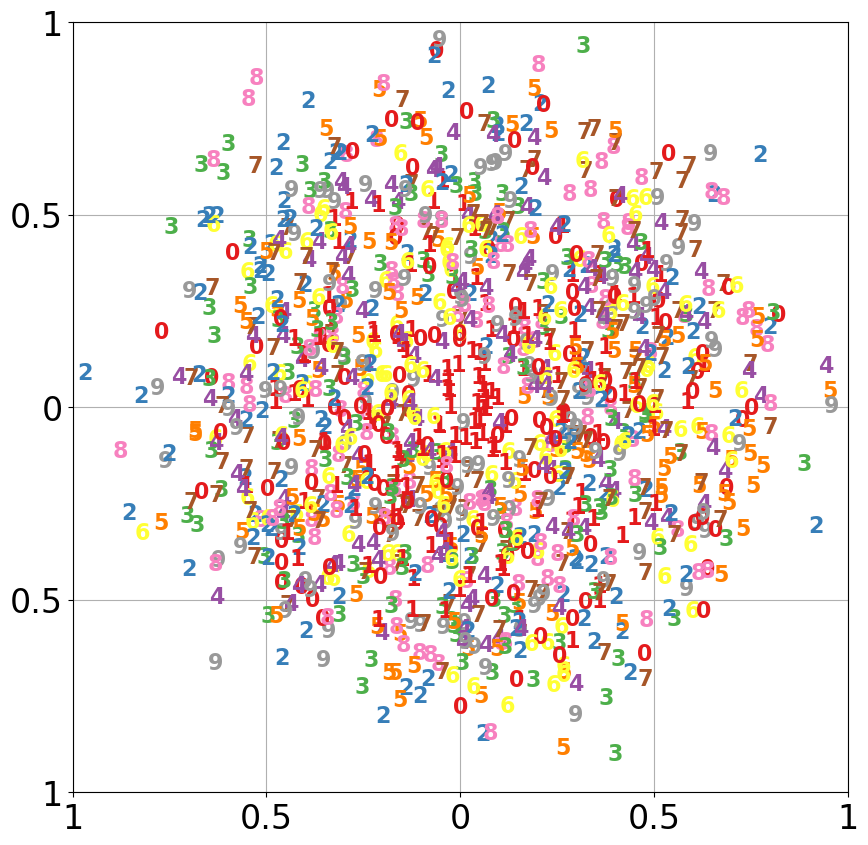} & \includegraphics[width=0.175\linewidth]{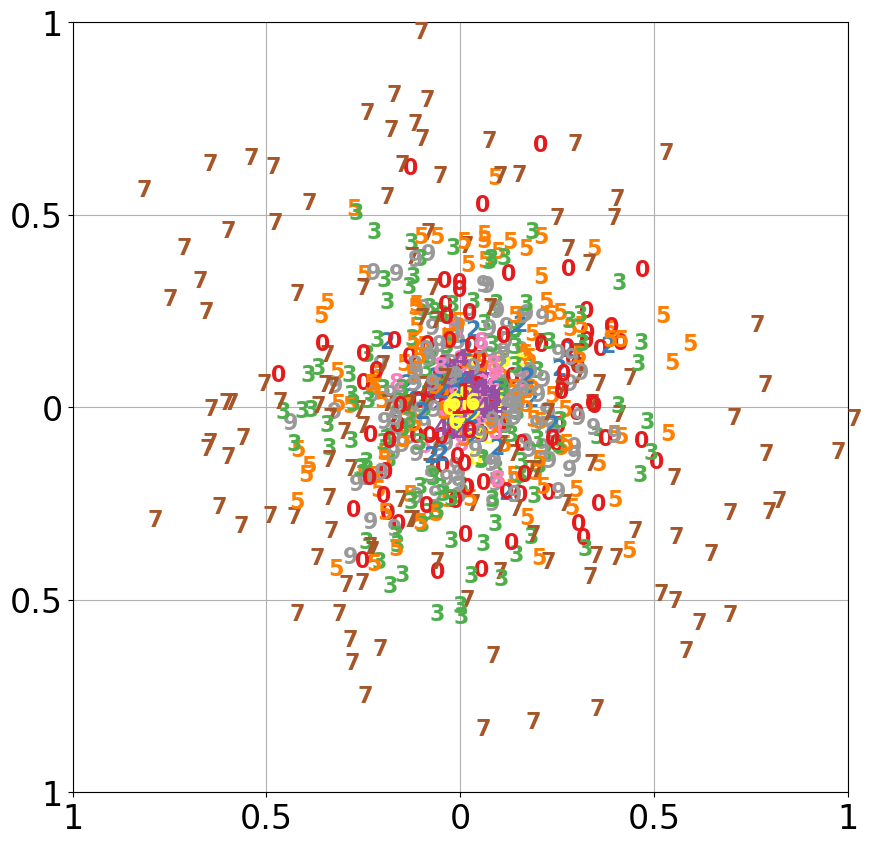} & \includegraphics[width=0.175\linewidth]{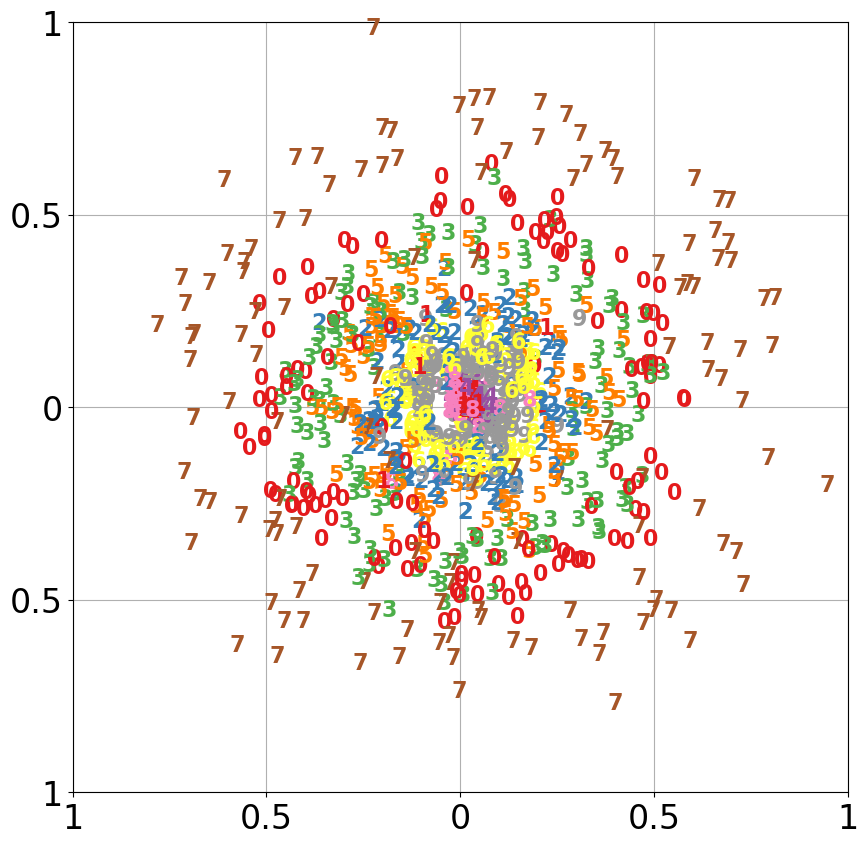} & \includegraphics[width=0.175\linewidth]{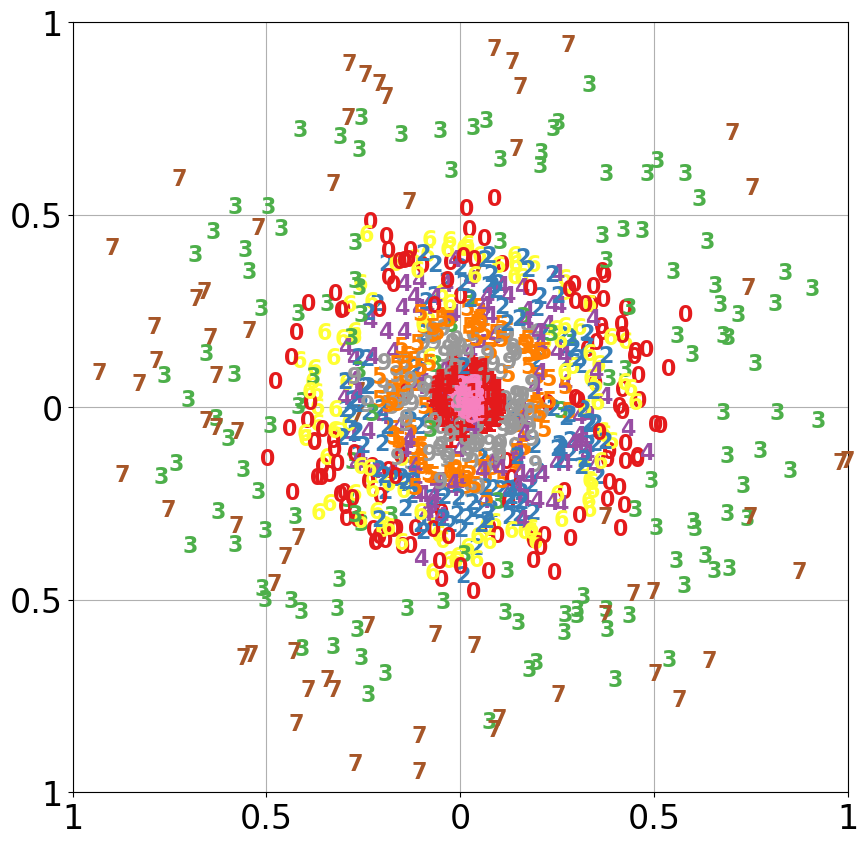} & \includegraphics[width=0.175\linewidth]{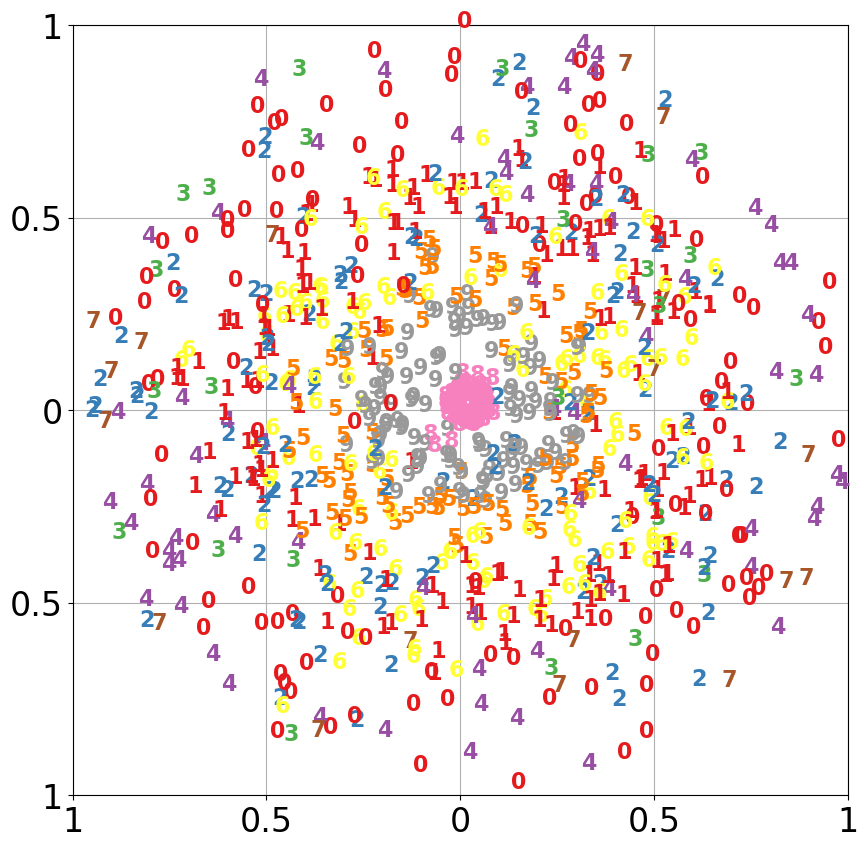} \\ \midrule
      
      \includegraphics[width=0.175\linewidth]{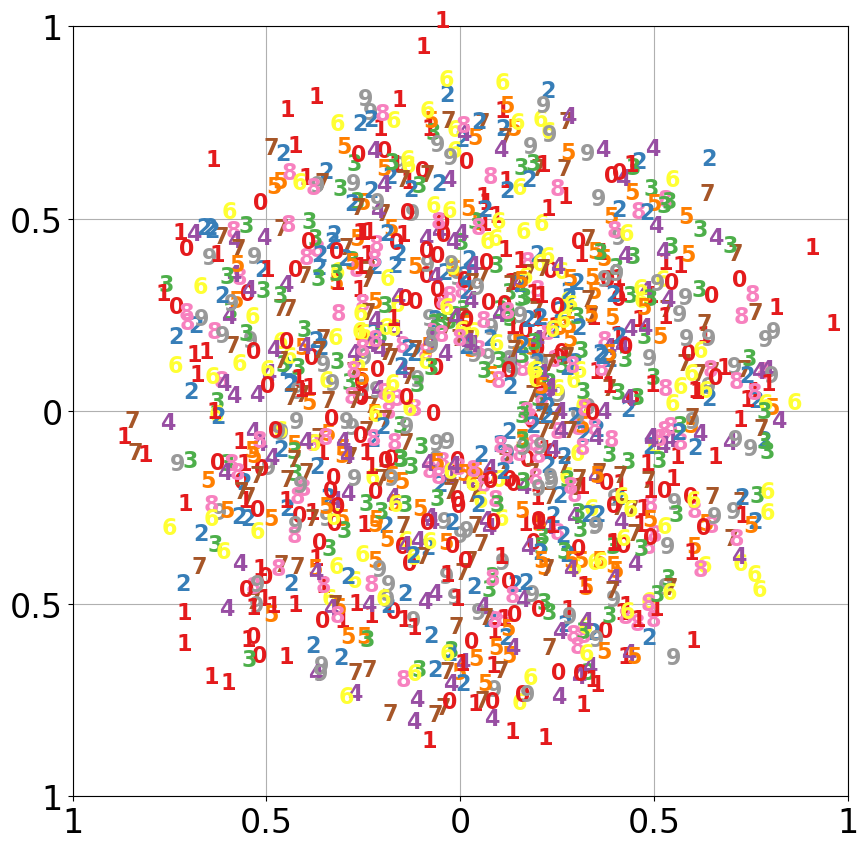} & \includegraphics[width=0.175\linewidth]{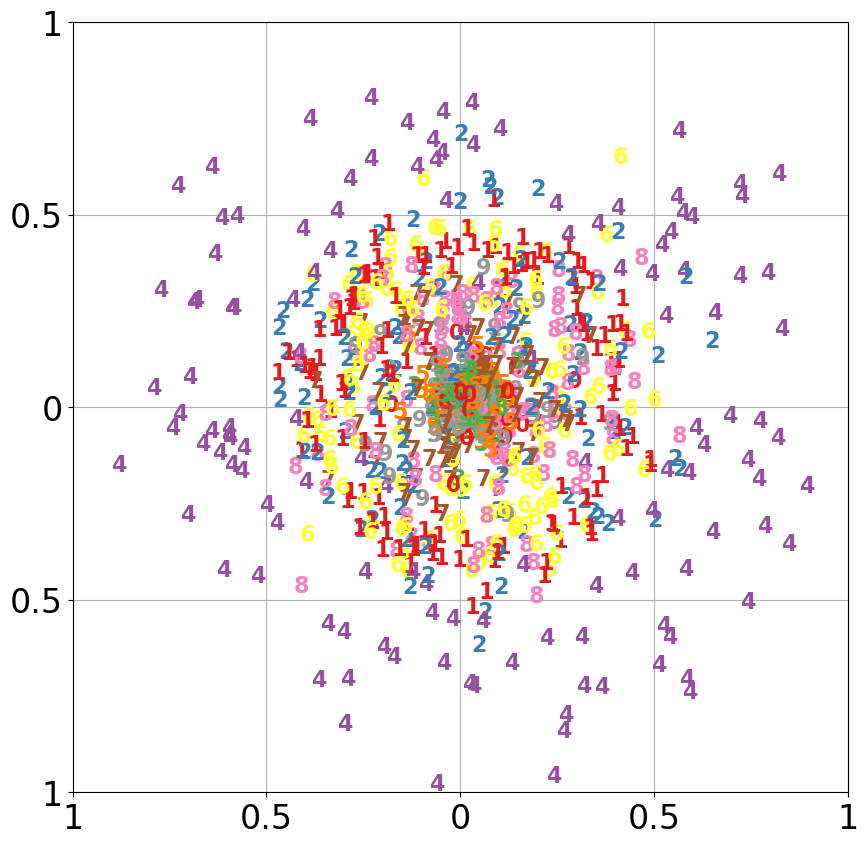} & \includegraphics[width=0.175\linewidth]{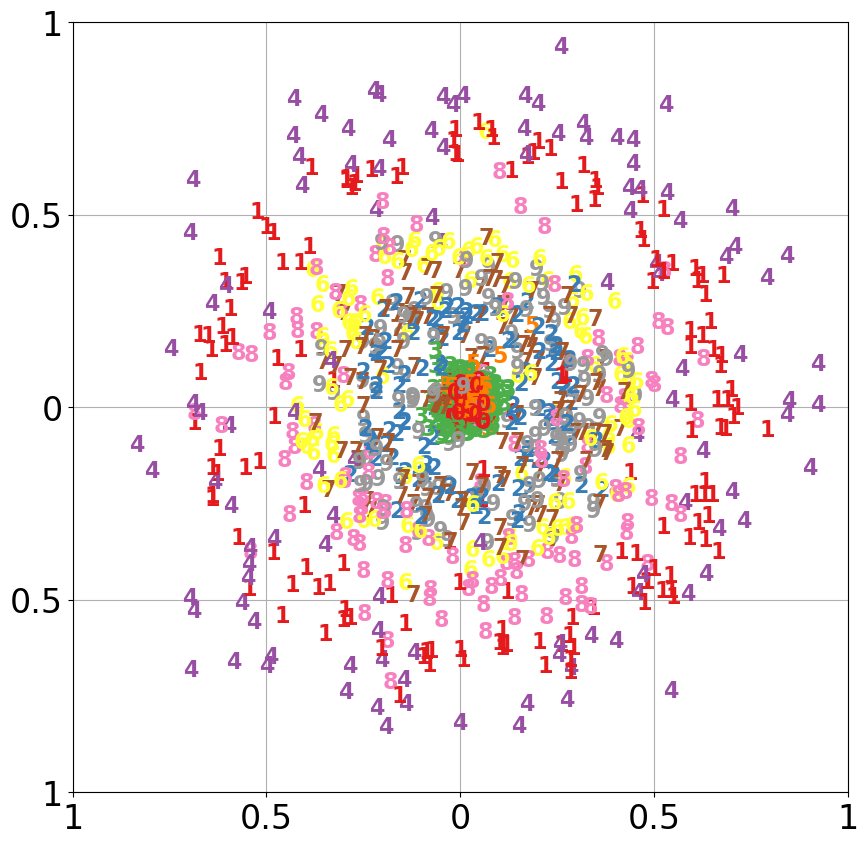} & \includegraphics[width=0.175\linewidth]{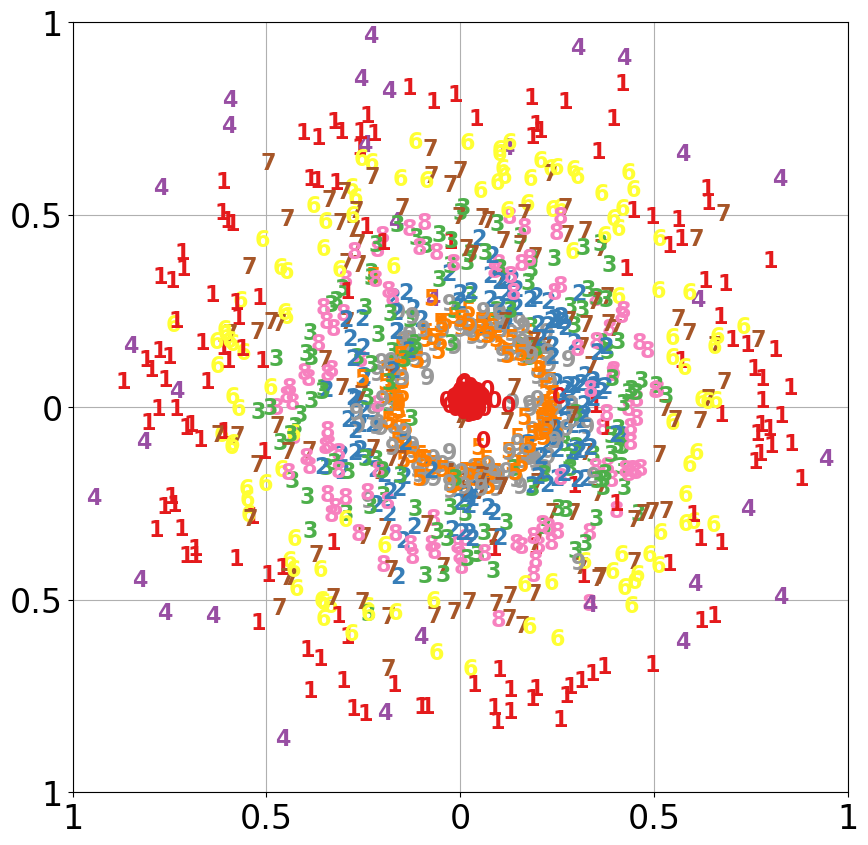} & \includegraphics[width=0.175\linewidth]{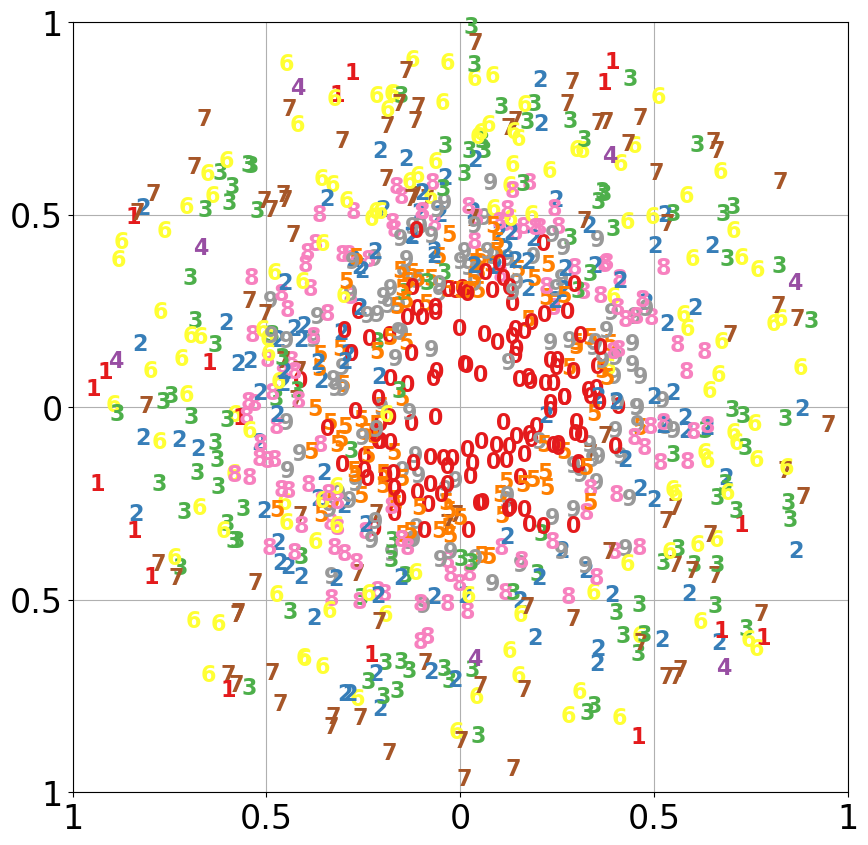} \\ \midrule 
      
      Before training & $50$\% of first epoch & First epoch & Third epoch & Fifth epoch  \\
       Test accuracy: $8.9$\% & Test accuracy: $20.3$\% & Test accuracy: $77.1$\% & Test accuracy: $97.5$\% & Test accuracy: $98.8$\%  \\
        
     \toprule
     \end{tabular}
     \end{center}
     \caption{The location of data samples compared to the cluster centers during the training process. The centers of the clusters are in the middle of the figures. The training samples are located by a random angle and their distance from the center of the cluster. The vertical and horizontal axes show the normalized distances.}
     \label{fig:single_prototype_in_training}
      
\end{table*}

\begin{table*}[!ht]
     \begin{center}
     \begin{tabular}{c c c c c}
     \toprule
     \includegraphics[width=0.175\linewidth]{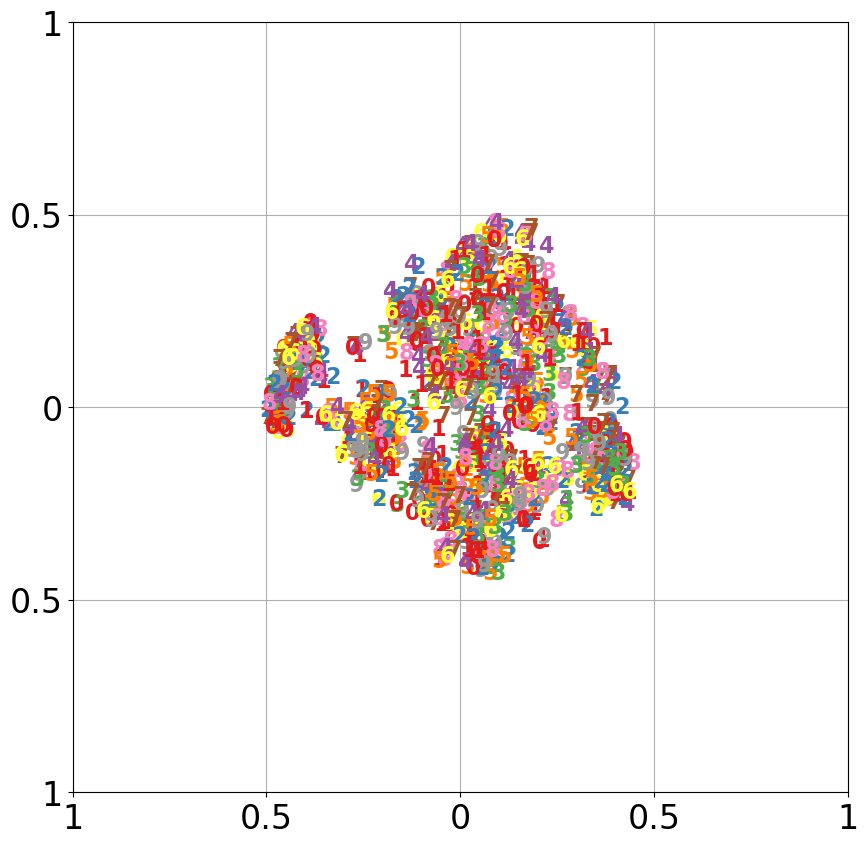} & \includegraphics[width=0.175\linewidth]{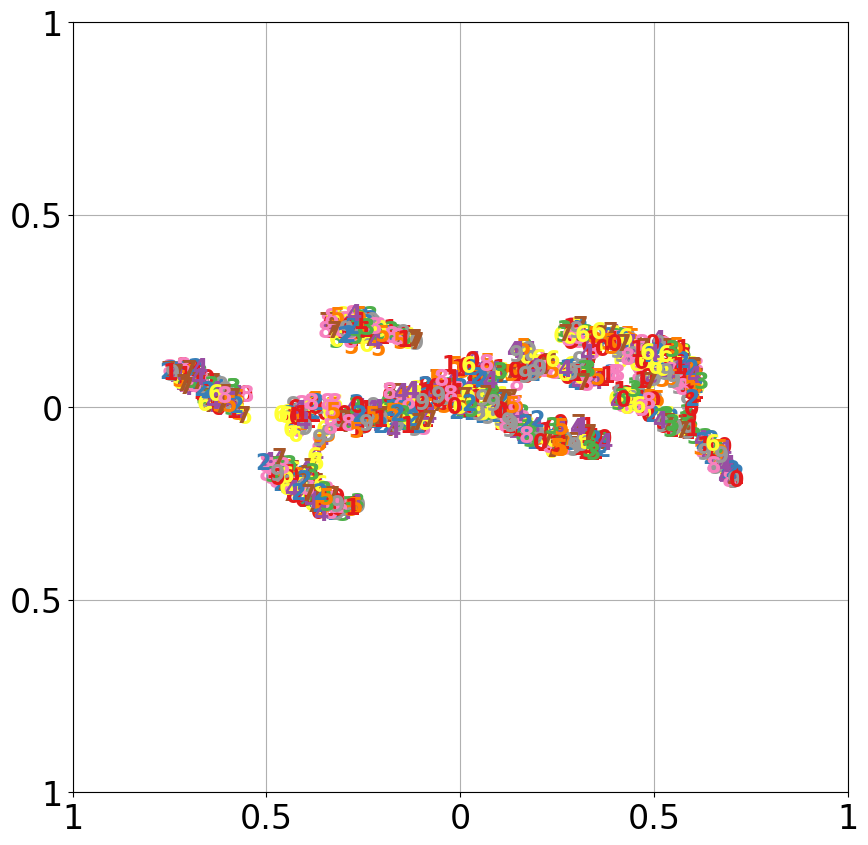} & \includegraphics[width=0.175\linewidth]{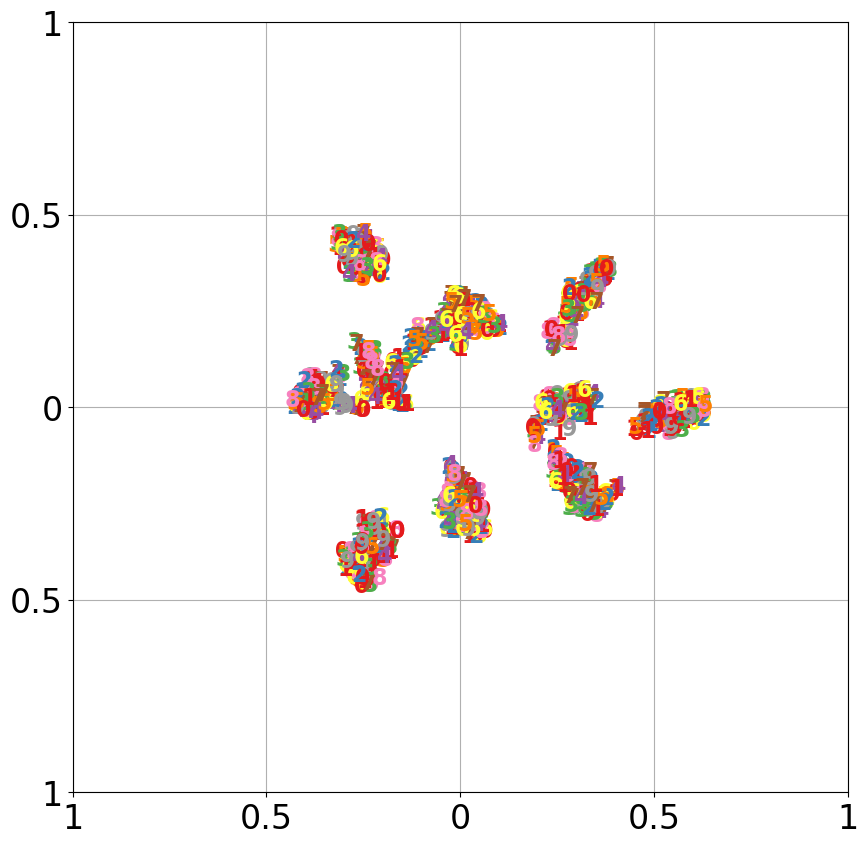} & \includegraphics[width=0.175\linewidth]{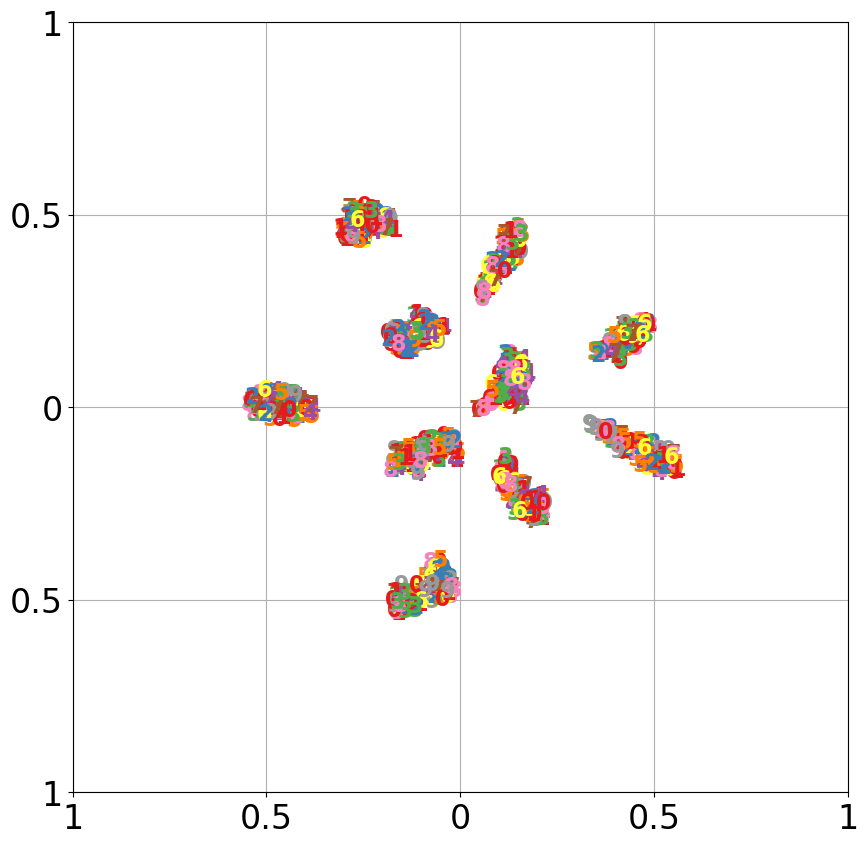} & \includegraphics[width=0.175\linewidth]{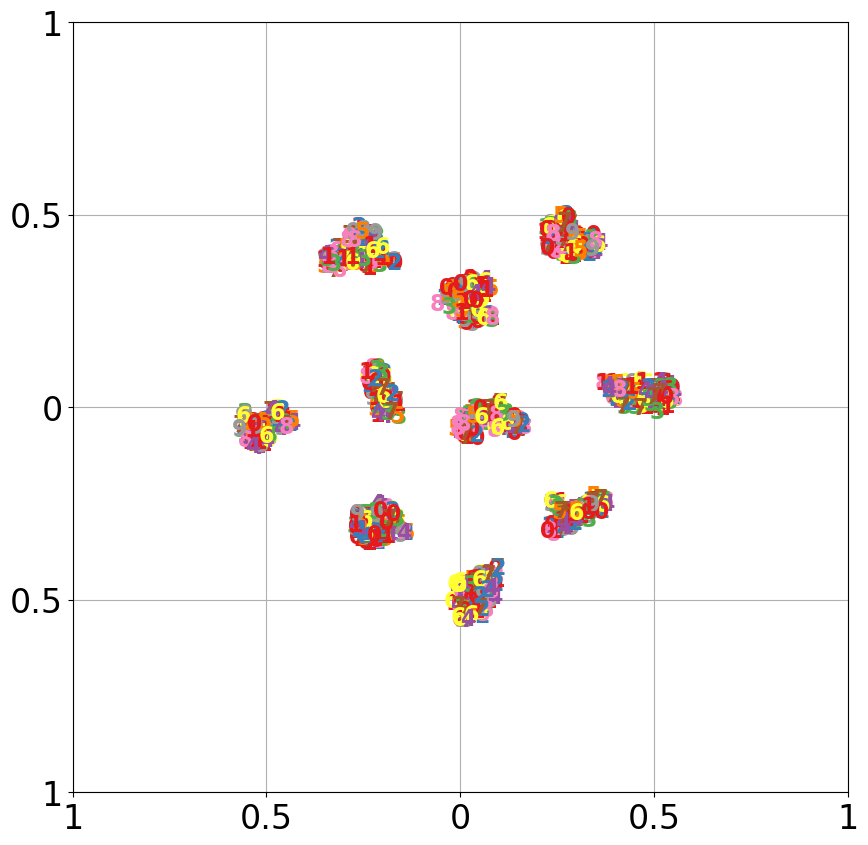} \\ \midrule
     \includegraphics[width=0.175\linewidth]{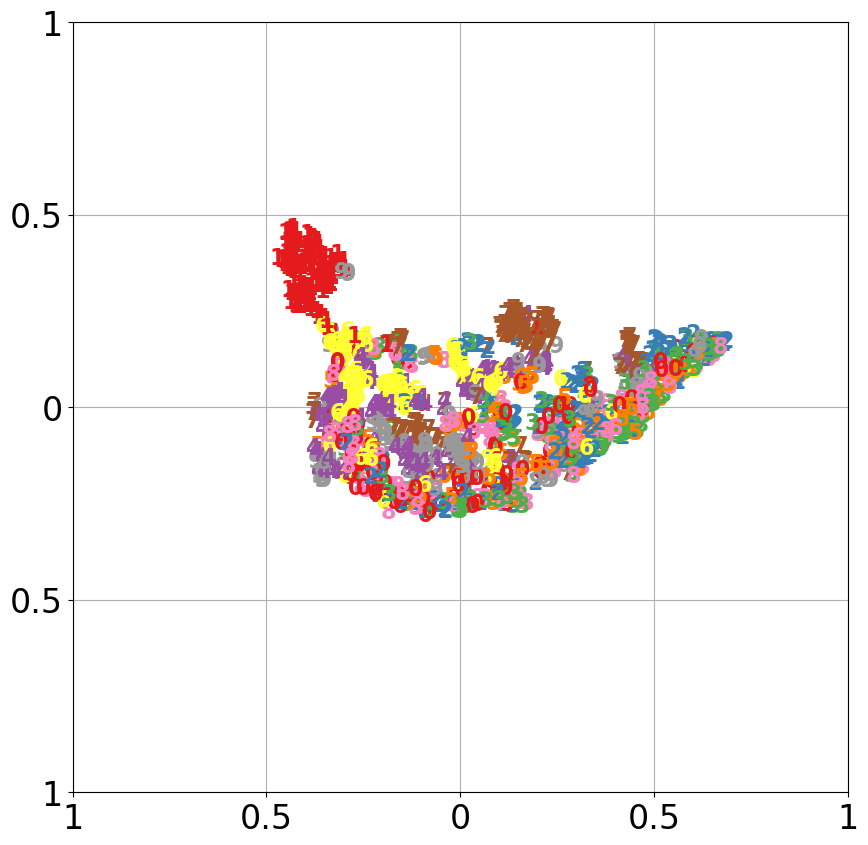} & \includegraphics[width=0.175\linewidth]{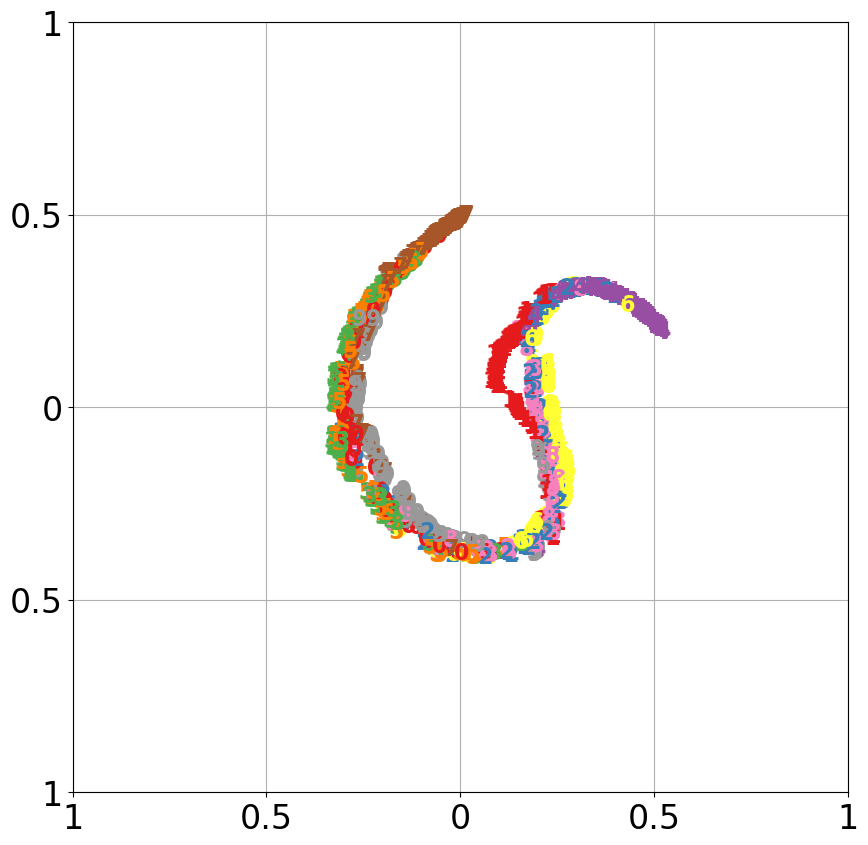} & \includegraphics[width=0.175\linewidth]{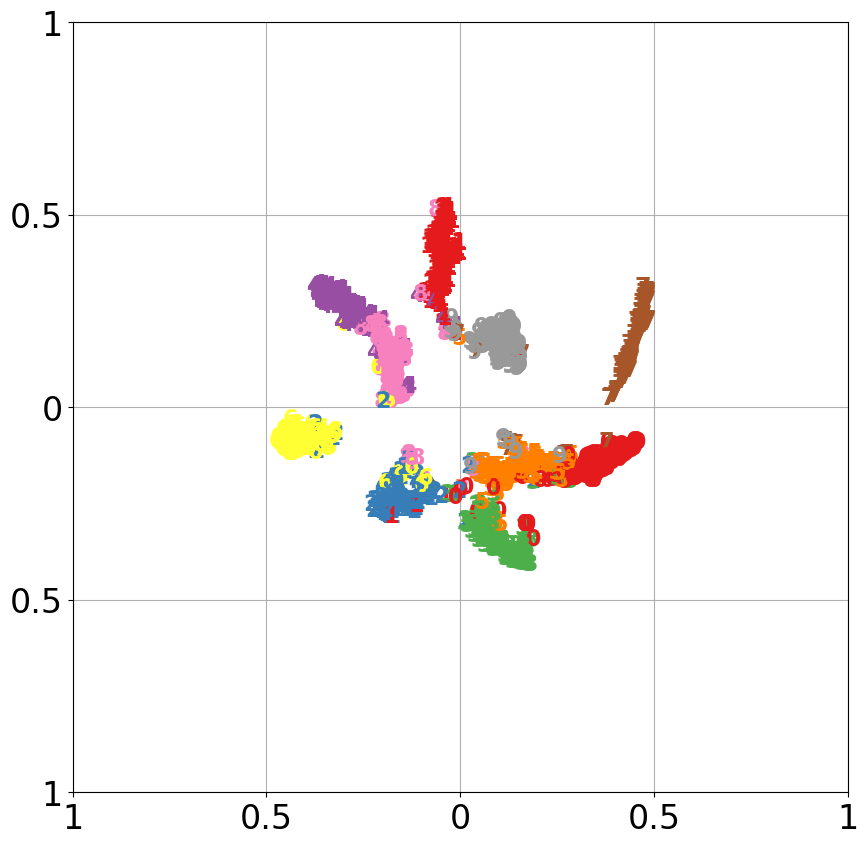} & \includegraphics[width=0.175\linewidth]{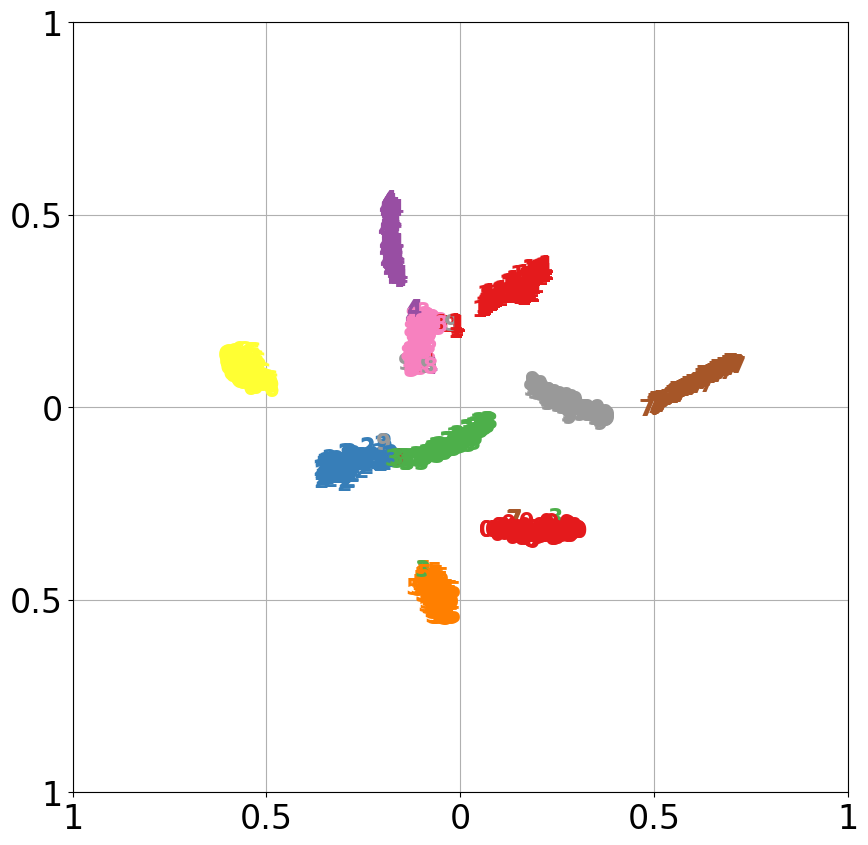} & \includegraphics[width=0.175\linewidth]{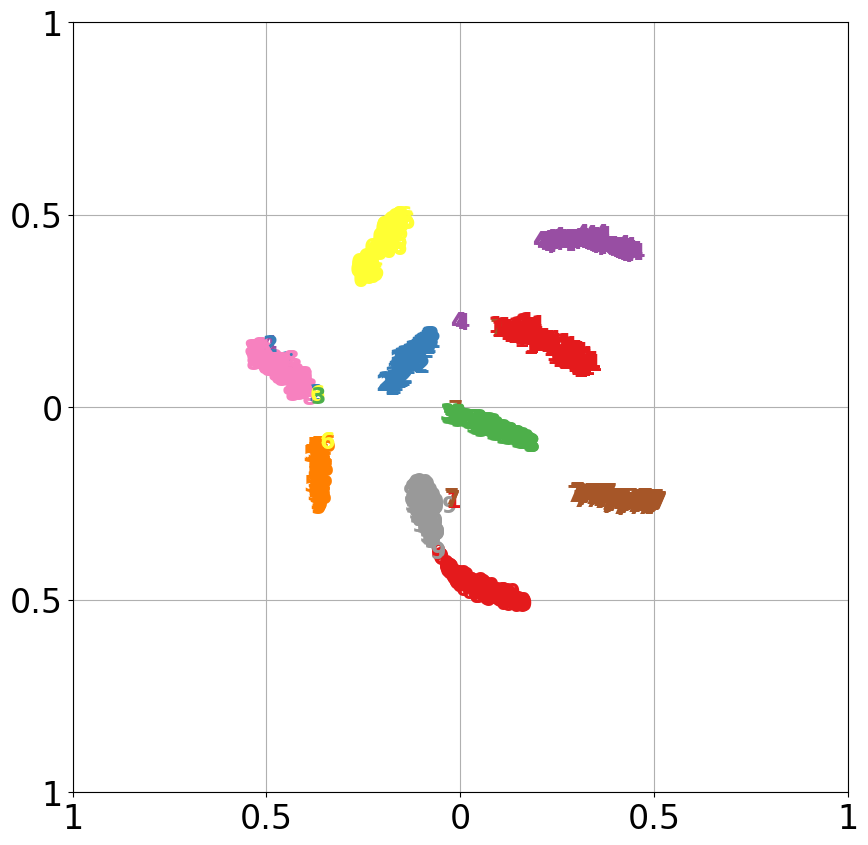} \\ \midrule      
     Before training & 50\% of first epoch & First epoch & Third epoch & Fifth epoch \\
     \toprule
     \end{tabular}
     \end{center}
     \caption{The two-dimensional representation of the training process. The figure presents the embeddings of the convolutional backbone (top row) and the activations of the RBFs (bottom row) and then maps them to a two-dimensional space using t-SNE \cite{van2012visualizing}. The vertical and horizontal axes depict the normalized values; however, the unit value is the same in all sub-figures.}
     \label{fig:tsne}
\end{table*}

It is conventional to use clustering algorithms such as the k-means or expectation-maximization (EM) algorithms to initialize the cluster centers. The loss function in Equation \ref{eq:LossRBF} is optimized using gradient descent by minimizing the distance of the embeddings for each sample from its nearest cluster center regardless of the class labels. The distance from the nearest cluster center is computed from the distance metric $\boldsymbol{R_j}$ \hl{defined in} Equation \ref{eq:distance}.

\subsection{Quadratic kernel}
 The kernels used for \hl{classical RBFs are nonlinear, increasing model complexity.} The architectures proposed in \ref{fig:CNN-RBF} profit from \hl{utilizing the state-of-the-art in representation learning, i.e. CNNs, as a backbone.} Therefore, CNN-RBF architectures can be trained with simpler linear models to improve the gradient flow \hl{during} backpropagation. The proposed quadratic activation function is linear in the space of $r^2$ and is defined as follows:
\begin{flalign}\label{eq:kernel_propose}
	h(r) = 1-r^2/\sigma^2  &&
\end{flalign}
where $\sigma$ is the parameter that determines the width of the kernel. The proposed kernel is depicted in Figure \ref{fig:activation} along with the conventional activation kernels. \hl{Our proposed quadratic kernel} reduces the nonlinearity of the CNN-RBF \hl{computational graph} for backpropagation. The squares of the distances between cluster centers and samples are computed by linear matrix multiplication in Equation \ref{eq:distance} and applying the proposed activation, which is linear for $r^2$. Thus, the gradients of \hl{the} deep embeddings propagate backwards through a distance computation with matrix multiplication and linear activations.
\section{Visualization of training process}
\label{sec:training}
In this section, we visualize the performance of the RBFs on top of CNNs on a simple dataset. The experiments are conducted on the modified national institute of standards and technology dataset (MNIST) \cite{lecun1998gradient}, which is a dataset of hand-written digits including 10 classes. Learning the dataset is considered a simple task in computer vision. The simplicity of the dataset and learning task provides us with the opportunity to visualize the training process at a fine level of detail. We chose the same number of cluster centers as classes (10) in the dataset to depict the training process of a CNN-RBF. The network architecture in this section consists of a four-layer CNN and the output of these layers is connected to the RBF after a global average pooling layer and a fully connected layer.

Figure \ref{fig:single_prototype_in_training} demonstrates the evolution of a representation around the cluster center during the training process. The data samples in Figure \ref{fig:single_prototype_in_training} are placed according to their distance from the center and at a random angle. The samples are shown with a number denoting their class and \hl{are additionally color code accordingly} in Figure \ref{fig:single_prototype_in_training}. To reduce the overlap of similar samples, we add random uniformly distributed noise at an amplitude of $0.1$ to the distance of the samples from \hl{the cluster} centers.

Minimizing the unsupervised loss in Equation \ref{eq:clus} reduces the distance of the data samples from the cluster centers. Furthermore, the supervised loss enforces the samples of the same class to \hl{maintain} the same distance from cluster centers, \hl{as} the activations are the only information for the final decision of the network. The circles with samples of the same class around the cluster centers demonstrate the effect of supervised loss in training. It should be noted that the clusters presented in Figure \ref{fig:single_prototype_in_training} are selected to optimally illustrate the concepts underlying training CNN-RBFs. 
Figure \ref{fig:tsne} illustrates the two-dimensional mapping of the CNN embeddings (top row) and RBF activations (bottom row) using t-SNE \cite{van2012visualizing}. The effect of both supervised and unsupervised loss from Equation \ref{eq:LossRBF} is \hl{also} visible in \hl{this} figure. The data samples split into clusters regardless of their class labels in the embedding space of CNN due to the unsupervised loss (top row in Figure \ref{fig:tsne}). The activation values divide into clusters corresponding to the class labels, a process encouraged by the supervised loss.

\section{Experimental Results}
\label{sec:experimental}
This section presents the experimental results that \hl{reinforce} the applicability of RBFs to CNNs. We use several standard computer vision benchmark datasets and investigate the effect of \hl{tweaking} various hyperparameters of \hl{the} CNN-RBF architectures in the training phase and generalization to test data. We used three convolutional backbones\hl{:} those of EfficientNet-B0 \cite{tan2019efficientnet}, InceptionV2 \cite{szegedy2016rethinking}, and ResNet50 \cite{he2016deep}. A list of the benchmark computer vision datasets is presented in Table \ref{table:datasets}.

\renewcommand{\tablename}{Table}
\setcounter{table}{0}
\begin{table}[!htb]
    \centering
    \begin{tabular}{@{}lrrr@{}}
        \toprule
        Dataset                                     & Train Size & Test Size & \# Classes \\ \midrule
        CIFAR-10 \cite{krizhevsky2009learning}      & 50 000     & 10 000    & 10         \\
        CIFAR-100 \cite{krizhevsky2009learning}     & 50 000     & 10 000    & 100        \\
        Oxford-IIIT Pets \cite{parkhi2012cats}      & 3 680      & 3 369     & 37         \\
        Oxford Flowers \cite{nilsback2008automated} & 1 020      & 6 140     & 102        \\
        FGVC Aircraft \cite{maji2013fine}           & 6 667      & 3 333     & 100        \\
        Caltech Birds \cite{WelinderEtal2010}       & 5 996      & 5 794     & 200        \\ \bottomrule
    \end{tabular}
    \caption{Computer vision benchmark datasets used to evaluate the performance of CNN-RBFs.}
    \label{table:datasets}
\end{table}

Figure \ref{fig:hyperparam} shows the hyperparameter search results for object \hl{classification} on two benchmark computer vision datasets: CIFAR-10 and CIFAR-100. The backbone CNN model in this experiment is EfficientNet-B0 \hl{with} RBFs on top of this backbone for classification. The image preprocessing pipeline, called AutoAugment \cite{cubuk2019autoaugment}, consists of a set of optimal and automatically searched augmentation policies for the ImageNet \cite{deng2009imagenet} dataset. The CNN-RBF architecture demonstrated in Figure \ref{fig:CNN-RBF} has two \hl{further} hyperparameters\hl{:} the number of cluster centers and the input dimensions of the RBF network. The models are optimized using an AdamW \cite{loshchilov2018decoupled} optimizer with learning rate and weight decay as hyperparameters. The loss constant $\lambda$ from Equation \ref{eq:LossRBF}, dropout rate, and batch size are the other hyperparameters.

The hyperparameter searches in Figure \ref{fig:hyperparam} are conducted using the hyperband \cite{li2017hyperband} algorithm with $4$ agents running in parallel on two \textit{Quadro T2000} graphic processing units (GPUs) for approximately $10$ days. It should be noted that dropout is only applied after the CNN backbone and before the fully connected layer in Figure \ref{fig:CNN-RBF}. The output of the fully connected layer, without any activation function, is used as the input feature of the RBFs. The results in Figure \ref{fig:hyperparam} shows that training CNN-RBF architectures leads to high performances with a \hl{wide} range of hyperparameters. However, achieving good test performance with a high dropout rate and a large input dimension is challenging. CNN-RBF architectures show a better performance without dropout and rectified linear unit (ReLU) activations in the input layer of RBFs. Thus, we neglect the dropout for further hyperparameter searches conducted on the datasets in Table \ref{table:datasets}. The list of optimal hyperparameters for all datasets is presented in Table \ref{table:list_of_hyperparameters}. 

\renewcommand{\tablename}{Figure}
\setcounter{table}{4}
\begin{table}[htb]
    \begin{center}
    \begin{tabular}{c }
    \toprule
    \includegraphics[width=0.45\textwidth]{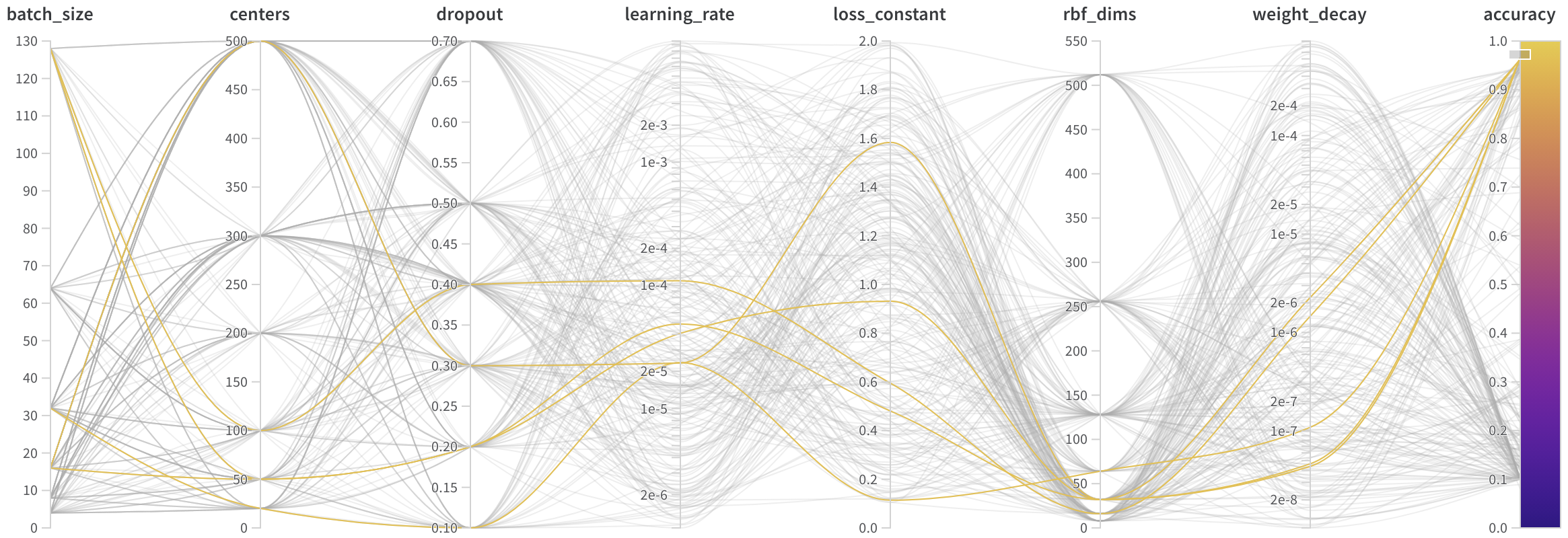}   \\ \midrule
    \includegraphics[width=0.45\textwidth]{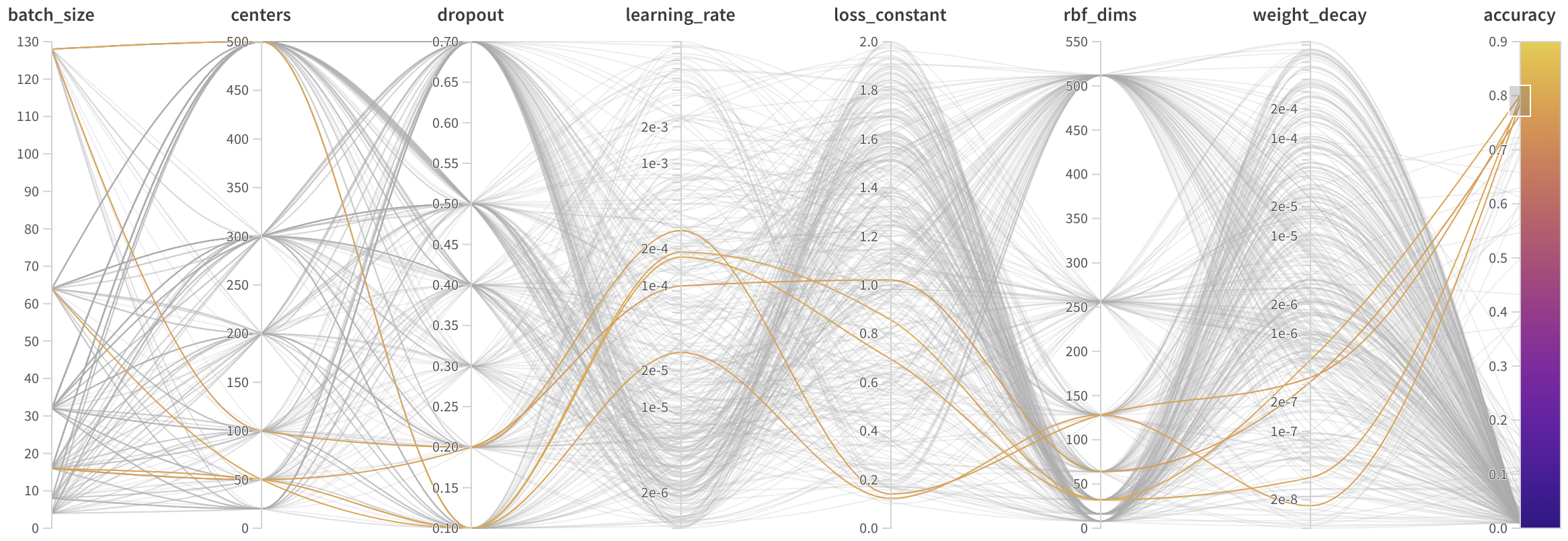} \\ 
    \toprule
    \end{tabular}
    \end{center}
    \caption{Hyperparameter search results from CIFAR-10 (top) and CIFAR-100 (bottom). The top five performing sets of hyperparameters for each dataset are highlighted in yellow.}
    \label{fig:hyperparam}
\end{table}

\hl{We also applied} CNN-RBFs to several other computer vision datasets with various backbone architectures. The experimental results of applying the CNN-RBFs to the computer vision benchmark datasets with the standard train and test splits are presented in Table \ref{table:efficientnet_dataset}. CNN-RBFs show the capacity to learn the entire dataset in all of the cases\hl{. There is, however,} a small gap between the best reported performances in computer vision literature and CNN-RBF architectures. Using dropout with CNN-RBFs for regularization does not lead to desirable results and reducing the number of parameters of \hl{the} RBFs\hl{,} while limiting the input size\hl{,} is the best regularization strategy that we found for RBFs besides data augmentation. Developing regularization methods for RBFs to improve generalization is an open research topic for reducing the gap between \hl{our} current results and \hl{those} of state-of-the-art computer vision models. 

\renewcommand{\tablename}{Table}
\setcounter{table}{1}
\begin{table}[htb]
    \resizebox{\columnwidth}{!}{
        \centering
    \begin{tabular}{l | c c c c c c}
     \toprule
      & Loss & Learning & Embeddings & Batch & Number of & Weight \\
     Dataset & constant & rate & dimensions & size & centers & decay \\
     \midrule
     CIFAR-10           & 0.1141 &  2.355e-5  &  64  &  32  &  20  &  1.090e-7 \\
     CIFAR-100          & 0.8557 &  1.873e-4  &  32  &  64  &  50  &  5.369e-7 \\
     Oxford-IIIT Pets   & 1.067  &  7.487e-5  &  64  &  16  &  50  &  1.150e-7 \\
     Oxford Flowers     & 1.562  &  1.076e-4  &  16  &  64  & 100  &  3.843e-6  \\     
     FGVC Aircraft      & 0.5471 &  1.103e-4  &  8   &   8  &  50  &  1.222e-6  \\
     Caltech Birds      & 0.5156 &  2.603e-4  &  32  &  32  &  50  &  1.416e-8  \\
    \toprule
    \end{tabular}}
    \caption{List of the final hyperparameters used for each computer vision benchmark dataset to achieve the performance of CNN-RBF architectures.}
    \label{table:list_of_hyperparameters}
\end{table}

\renewcommand{\tablename}{Table}
\setcounter{table}{2}

\begin{table}[t!]
\begin{center}
\resizebox{0.45\textwidth}{!}{  
\begin{tabular}{ l | c | c c c | c }
\toprule
\multirow{2}{*}{Dataset} & & \multicolumn{3}{c|}{CNN-RBFs} & Best \\ 
& Backbone & EfficientNet-B0 & InceptionV2 & ResNet50 & result \\ 
\midrule
\multirow{2}{*}{CIFAR-10} & No-Augment & 0.966 & 0.963 & 0.969 & 0.993   \\
& Auto-Augment  & 0.975 & \textbf{0.977} & 0.942    \\ \midrule

\multirow{2}{*}{CIFAR-100} & No-Augment & 0.797 & 0.752 & 0.693 & 0.936   \\
& Auto-Augment & \textbf{0.822} & 0.805 & 0.778   \\ \midrule

\multirow{2}{*}{Oxford-IIIT Pets} & No-Augment  & 0.840 & 0.804 & 0.622 & 0.967   \\
& Auto-Augment  & \textbf{0.887} & 0.820 & 0.829  \\ \midrule

\multirow{2}{*}{Oxford Flowers} & No-Augment & 0.609 & 0.659 & 0.595 & 0.997   \\
& Auto-Augment  & \textbf{0.828} & 0.757 & 0.667  \\ \midrule

\multirow{2}{*}{FGVC Aircraft} & No-Augment & 0.723 & 0.717 & 0.665 & 0.945  \\
& Auto-Augment & 0.842 & \textbf{0.843} & 0.828 \\ \midrule

\multirow{2}{*}{Caltech Birds} & No-Augment  & 0.613 & 0.428 & 0.281 & 0.904  \\
& Auto-Augment  & \textbf{0.618} & 0.587 & 0.503  	 \\ \midrule

\end{tabular}}
\end{center}
\caption{Comparing the performance of various CNN-RBF architectures with pretraining and augmentation on benchmark computer vision datasets. \hl{The best results column is the top performance of the current state-of-the-art architecture on the benchmark dataset.}}
\label{table:efficientnet_dataset}
\end{table}

\section{Similarity Metric Learning and Interpretability of CNN-RBFs}
\label{sec:interpret}

\renewcommand{\tablename}{Figure}
\setcounter{table}{5}
\begin{table*}[!ht]
    \centering
     \resizebox{0.87\textwidth}{!}{
     \begin{tabular}{c | c c c c c c c}
     \toprule
     Test image & \multicolumn{7}{c}{Similar and dissimilar images} \\
     \midrule
     \includegraphics[width=2cm]{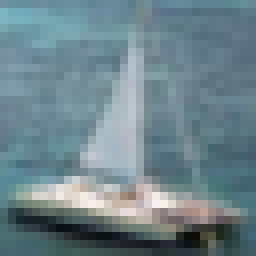} &    
     \includegraphics[width=2cm]{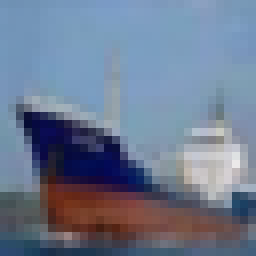} &
     \includegraphics[width=2cm]{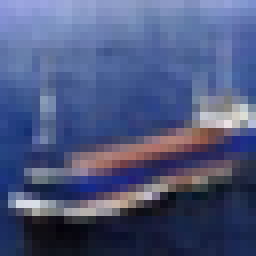} &
     \includegraphics[width=2cm]{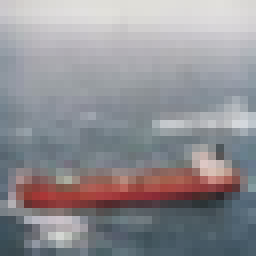} &
     \includegraphics[width=2cm]{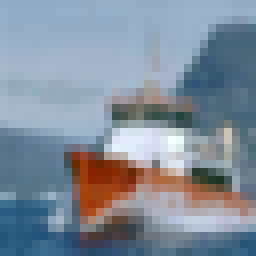} &
     \includegraphics[width=2cm]{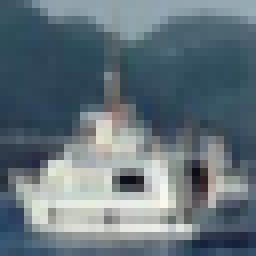} &
     \includegraphics[width=2cm]{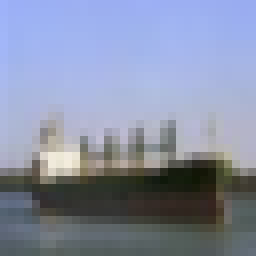} &
     \includegraphics[width=2cm]{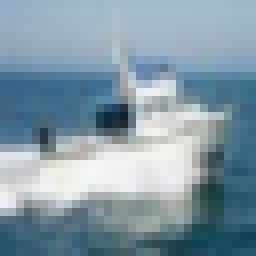} \\
     \midrule
     \includegraphics[width=2cm]{images/interpret_metric/metric/test_sample_1.png} &    
     \includegraphics[width=2cm]{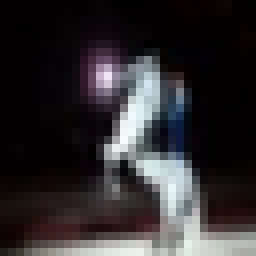} &
     \includegraphics[width=2cm]{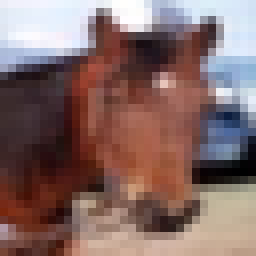} &
     \includegraphics[width=2cm]{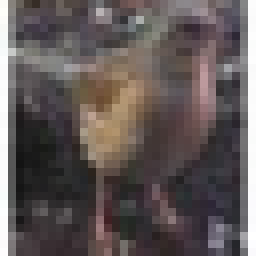} &
     \includegraphics[width=2cm]{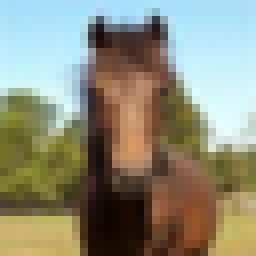} &
     \includegraphics[width=2cm]{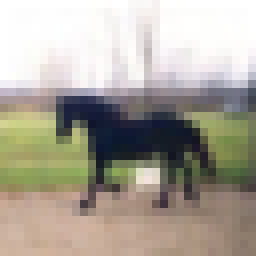} &
     \includegraphics[width=2cm]{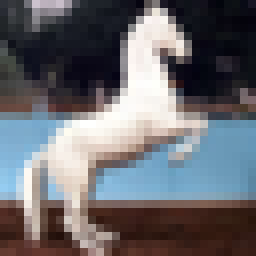} &
     \includegraphics[width=2cm]{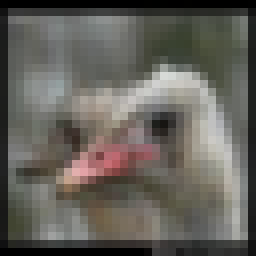} \\
     
     \toprule
     \includegraphics[width=2cm]{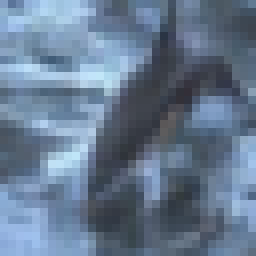} &    
     \includegraphics[width=2cm]{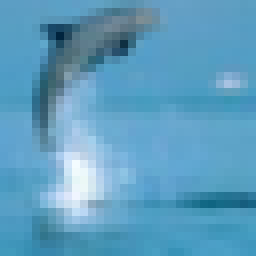} &
     \includegraphics[width=2cm]{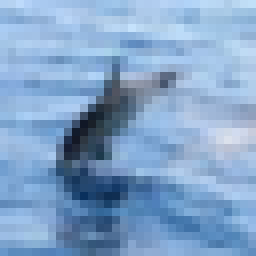} &
     \includegraphics[width=2cm]{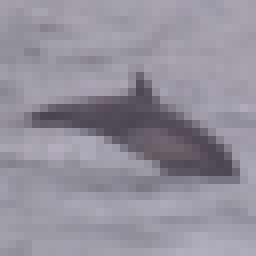} &
     \includegraphics[width=2cm]{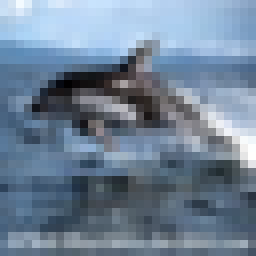} &
     \includegraphics[width=2cm]{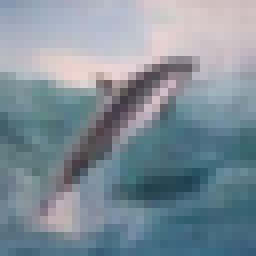} &
     \includegraphics[width=2cm]{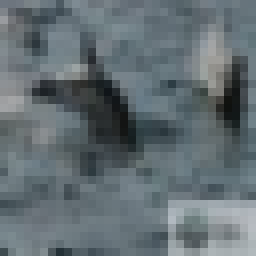} &
     \includegraphics[width=2cm]{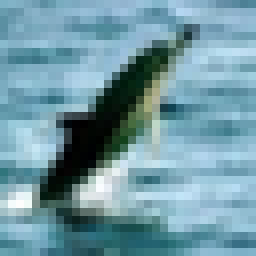} \\
     \midrule
    \includegraphics[width=2cm]{images/interpret_metric/metric/test_sample_2.png} &    
     \includegraphics[width=2cm]{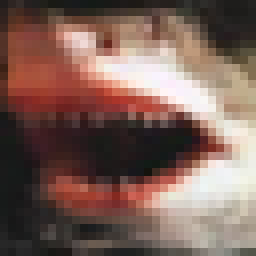} &
     \includegraphics[width=2cm]{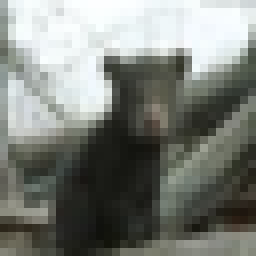} &
     \includegraphics[width=2cm]{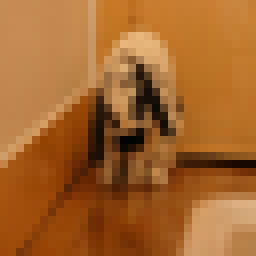} &
     \includegraphics[width=2cm]{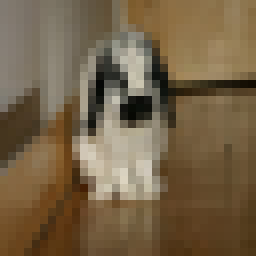} &
     \includegraphics[width=2cm]{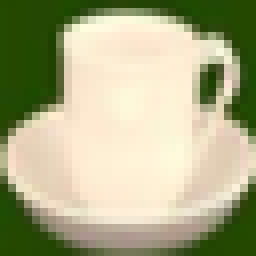} &
     \includegraphics[width=2cm]{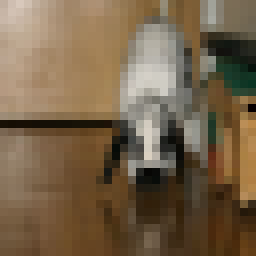} &
     \includegraphics[width=2cm]{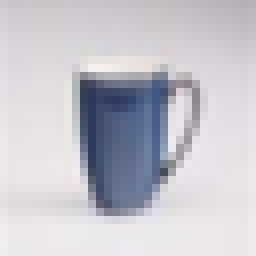} \\
     
     \toprule
     \includegraphics[width=2cm]{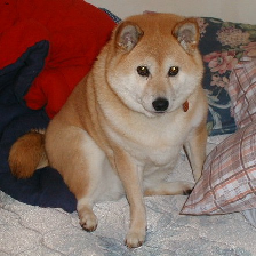} &    
     \includegraphics[width=2cm]{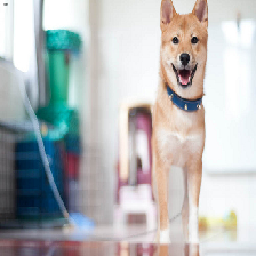} &
     \includegraphics[width=2cm]{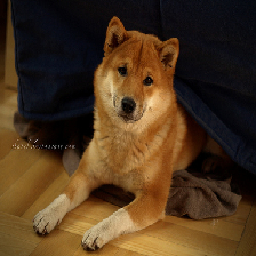} &
     \includegraphics[width=2cm]{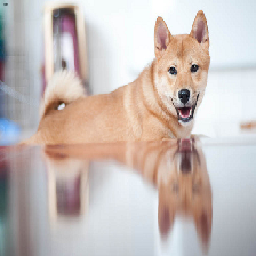} &
     \includegraphics[width=2cm]{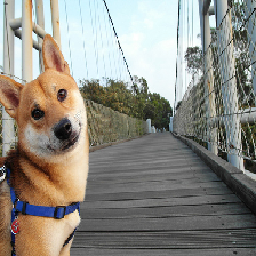} &
     \includegraphics[width=2cm]{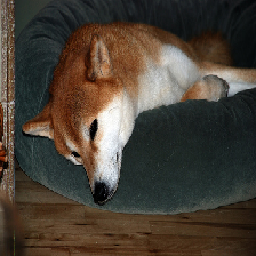} &
     \includegraphics[width=2cm]{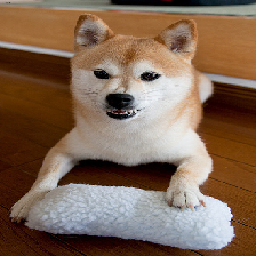} &
     \includegraphics[width=2cm]{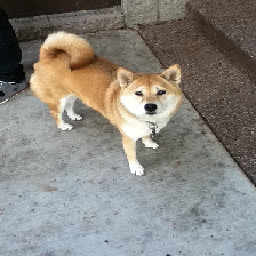} \\
     \midrule
    \includegraphics[width=2cm]{images/interpret_metric/metric/test_sample_3.png} &    
     \includegraphics[width=2cm]{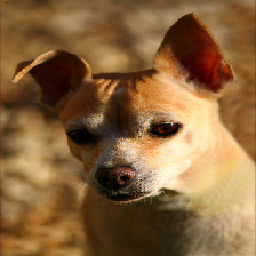} &
     \includegraphics[width=2cm]{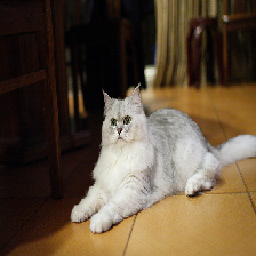} &
     \includegraphics[width=2cm]{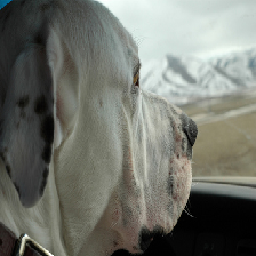} &
     \includegraphics[width=2cm]{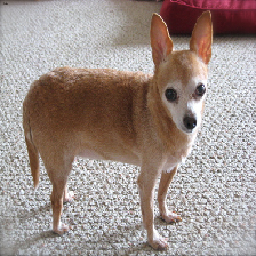} &
     \includegraphics[width=2cm]{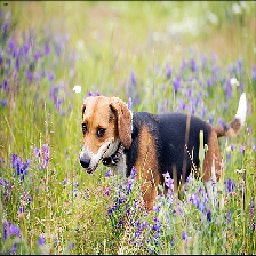} &
     \includegraphics[width=2cm]{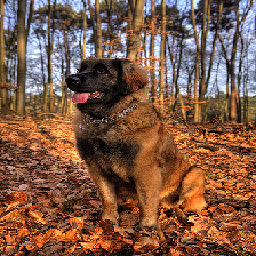} &
     \includegraphics[width=2cm]{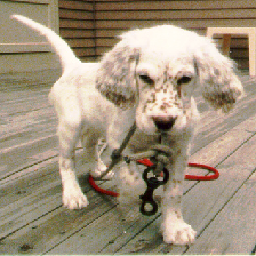} \\
     
     \toprule
     \includegraphics[width=2cm]{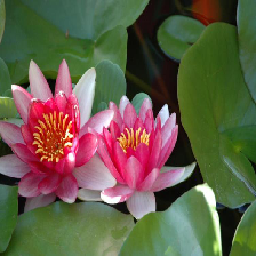} &    
     \includegraphics[width=2cm]{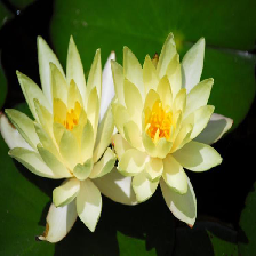} &
     \includegraphics[width=2cm]{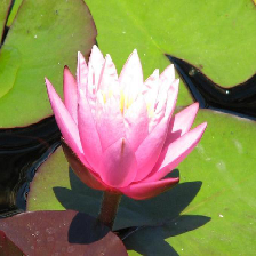} &
     \includegraphics[width=2cm]{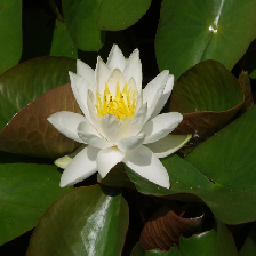} &
     \includegraphics[width=2cm]{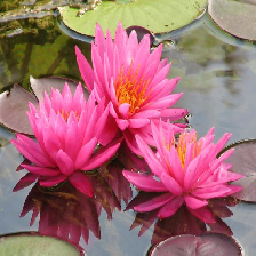} &
     \includegraphics[width=2cm]{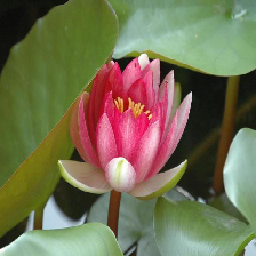} &
     \includegraphics[width=2cm]{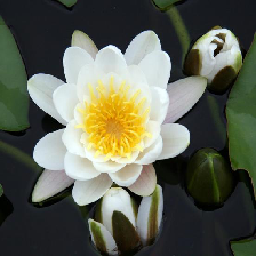} &
     \includegraphics[width=2cm]{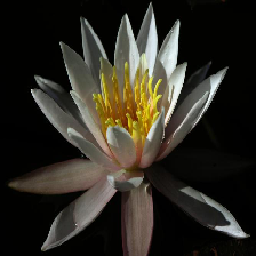} \\
     \midrule
    \includegraphics[width=2cm]{images/interpret_metric/metric/test_sample_4.png} &    
     \includegraphics[width=2cm]{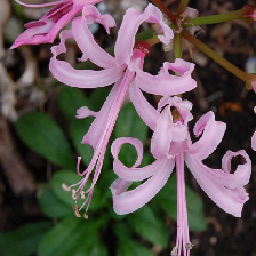} &
     \includegraphics[width=2cm]{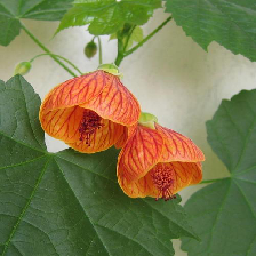} &
     \includegraphics[width=2cm]{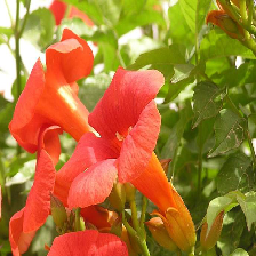} &
     \includegraphics[width=2cm]{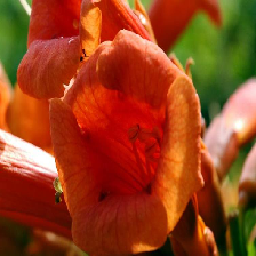} &
     \includegraphics[width=2cm]{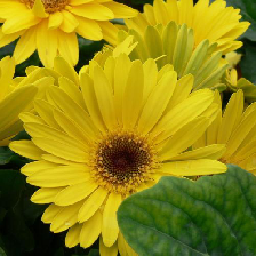} &
     \includegraphics[width=2cm]{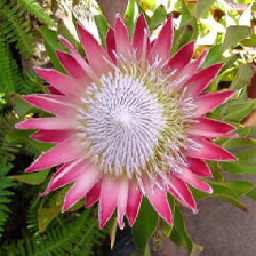} &
     \includegraphics[width=2cm]{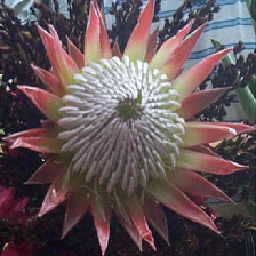} \\
     
     \toprule
     \includegraphics[width=2cm]{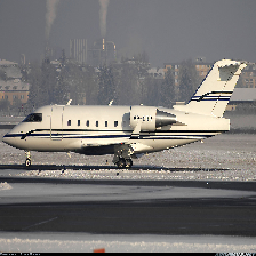} &    
     \includegraphics[width=2cm]{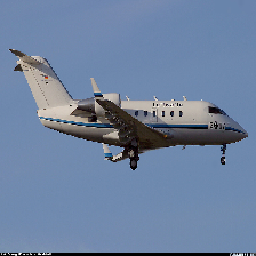} &
     \includegraphics[width=2cm]{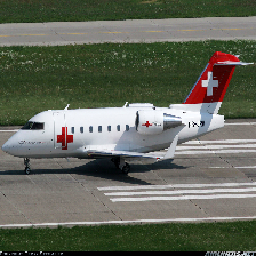} &
     \includegraphics[width=2cm]{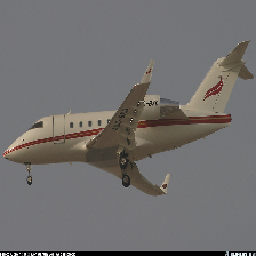} &
     \includegraphics[width=2cm]{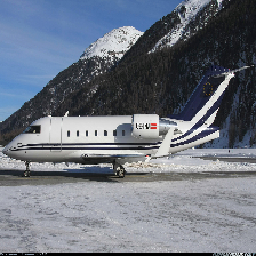} &
     \includegraphics[width=2cm]{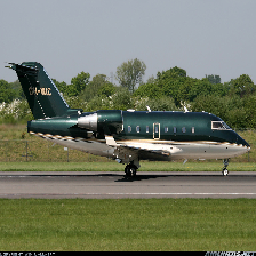} &
     \includegraphics[width=2cm]{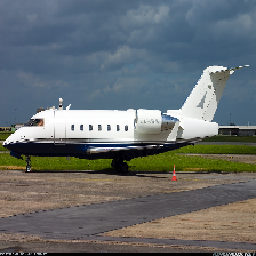} &
     \includegraphics[width=2cm]{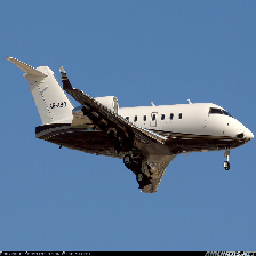} \\
     \midrule
    \includegraphics[width=2cm]{images/interpret_metric/metric/test_sample_5.png} &    
     \includegraphics[width=2cm]{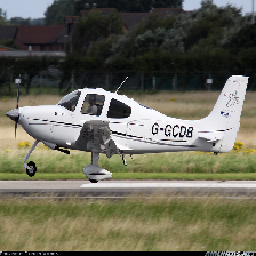} &
     \includegraphics[width=2cm]{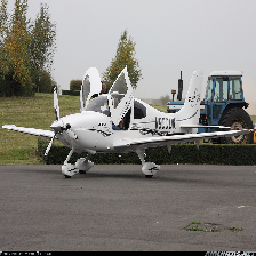} &
     \includegraphics[width=2cm]{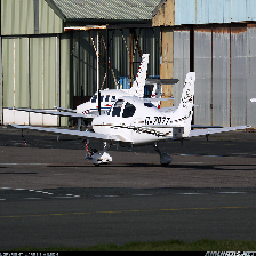} &
     \includegraphics[width=2cm]{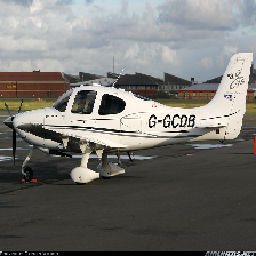} &
     \includegraphics[width=2cm]{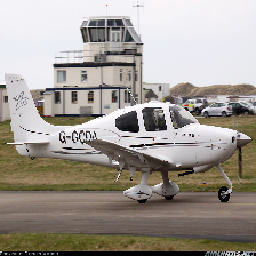} &
     \includegraphics[width=2cm]{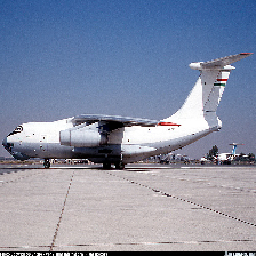} &
     \includegraphics[width=2cm]{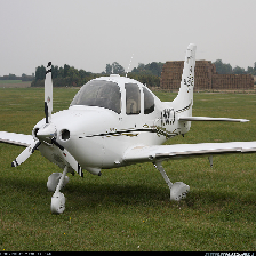} \\
     
     \toprule
     \includegraphics[width=2cm]{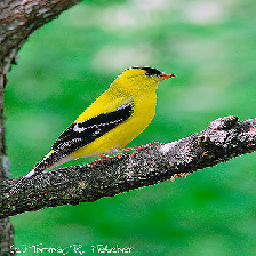} &    
     \includegraphics[width=2cm]{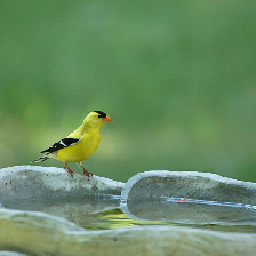} &
     \includegraphics[width=2cm]{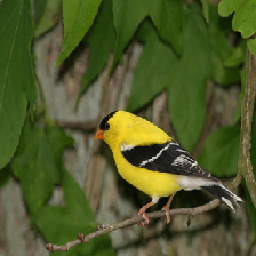} &
     \includegraphics[width=2cm]{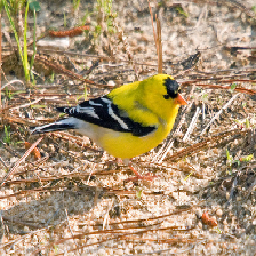} &
     \includegraphics[width=2cm]{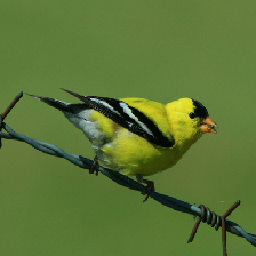} &
     \includegraphics[width=2cm]{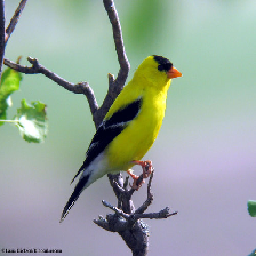} &
     \includegraphics[width=2cm]{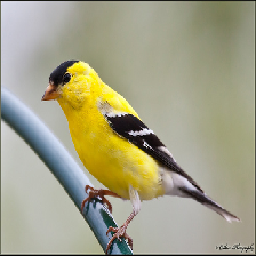} &
     \includegraphics[width=2cm]{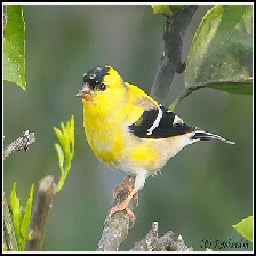} \\
     \midrule
    \includegraphics[width=2cm]{images/interpret_metric/metric/test_sample_6.png} &    
     \includegraphics[width=2cm]{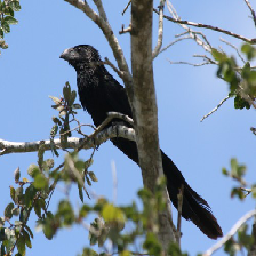} &
     \includegraphics[width=2cm]{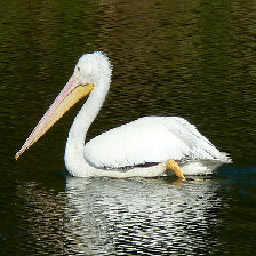} &
     \includegraphics[width=2cm]{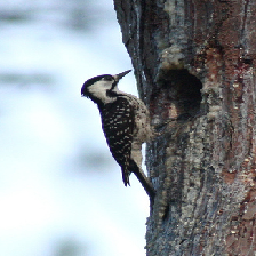} &
     \includegraphics[width=2cm]{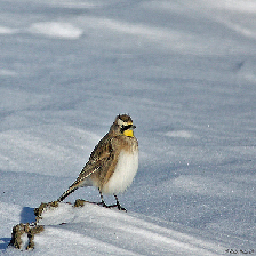} &
     \includegraphics[width=2cm]{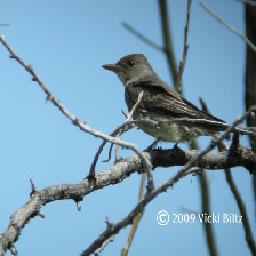} &
     \includegraphics[width=2cm]{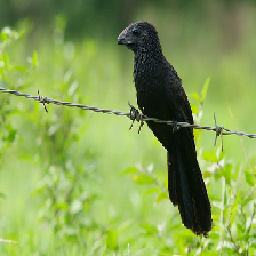} &
     \includegraphics[width=2cm]{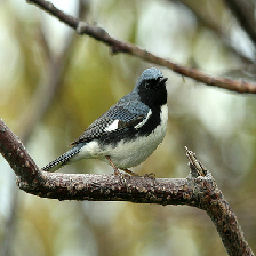} \\
     
     \toprule     
     \end{tabular}}
     \caption{Similar and dissimilar training images for given test images based on the similarity metric compute in Equation \ref{eq:distance}. The Figure depicts the top $7$ most similar and dissimilar training images for a given test image in every two rows. The images shown in every two consecutive rows belong to one of the datasets in Table \ref{table:datasets} with the same order.}
     \label{fig:similar_dissimilar}
\end{table*}

\begin{table*}[!ht]
    \centering
     \resizebox{0.87\textwidth}{!}{
     \begin{tabular}{c | c c c c c c c}
     \toprule
     Test image & \multicolumn{7}{c}{Similar images from training set in the embedding space} \\
     \midrule
     \includegraphics[width=2cm]{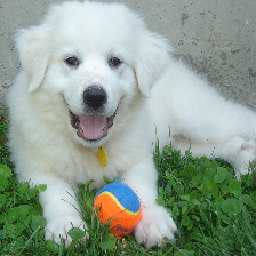} &    
     \includegraphics[width=2cm]{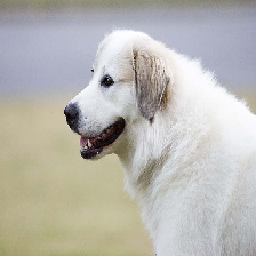} &
     \includegraphics[width=2cm]{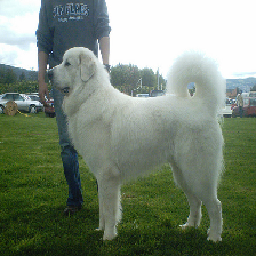} &
     \includegraphics[width=2cm]{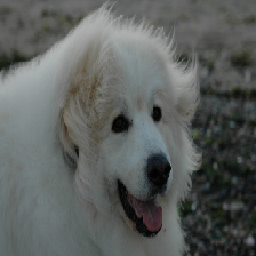} &
     \includegraphics[width=2cm]{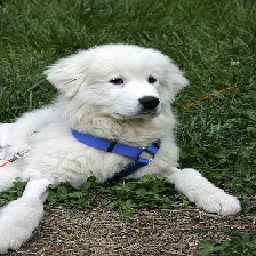} &
     \includegraphics[width=2cm]{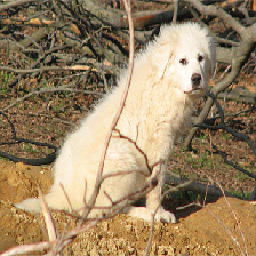} &
     \includegraphics[width=2cm]{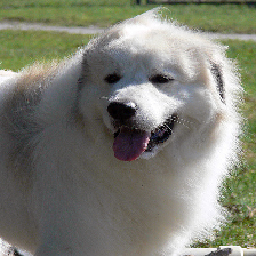} &
     \includegraphics[width=2cm]{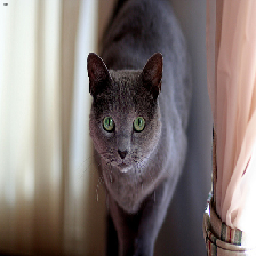} \\
     Learned metric &    
     \includegraphics[width=2cm]{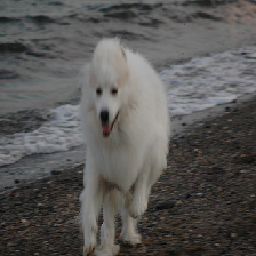} &
     \includegraphics[width=2cm]{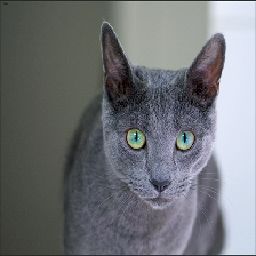} &
     \includegraphics[width=2cm]{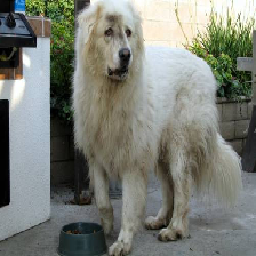} &
     \includegraphics[width=2cm]{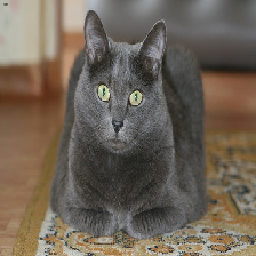} &
     \includegraphics[width=2cm]{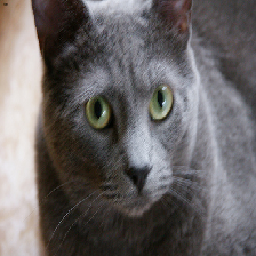} &
     \includegraphics[width=2cm]{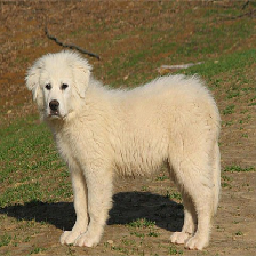} &
     \includegraphics[width=2cm]{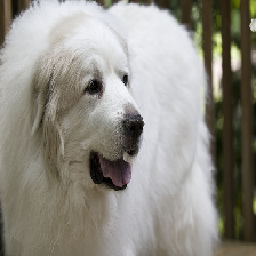} \\
     
     \toprule
     \includegraphics[width=2cm]{images/distances/test_image.png} &    
     \includegraphics[width=2cm]{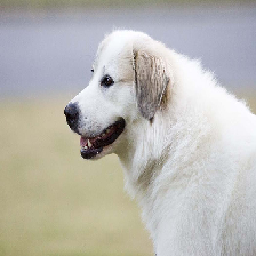} &
     \includegraphics[width=2cm]{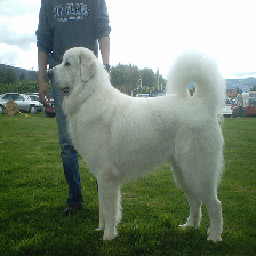} &
     \includegraphics[width=2cm]{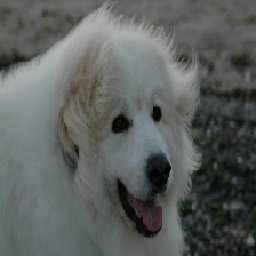} &
     \includegraphics[width=2cm]{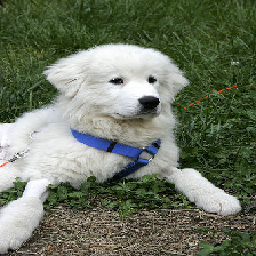} &
     \includegraphics[width=2cm]{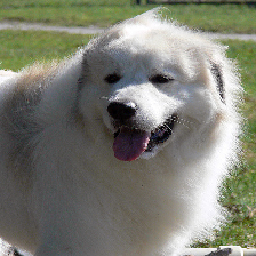} &
     \includegraphics[width=2cm]{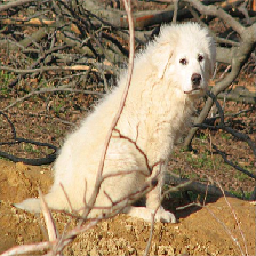} &
     \includegraphics[width=2cm]{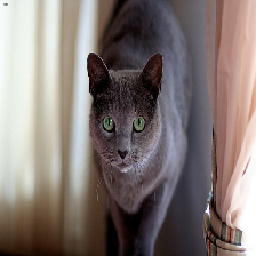} \\
     Euclidean distance &    
     \includegraphics[width=2cm]{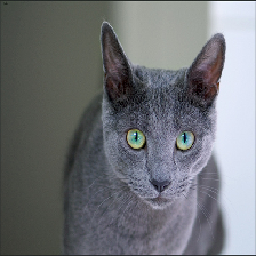} &
     \includegraphics[width=2cm]{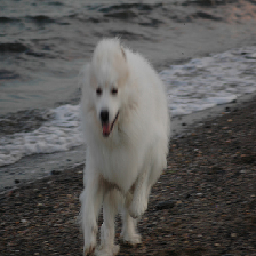} &
     \includegraphics[width=2cm]{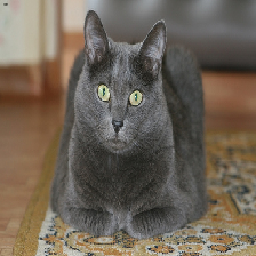} &
     \includegraphics[width=2cm]{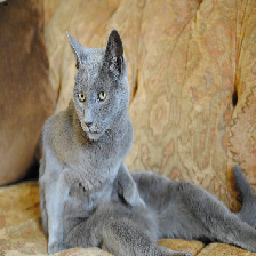} &
     \includegraphics[width=2cm]{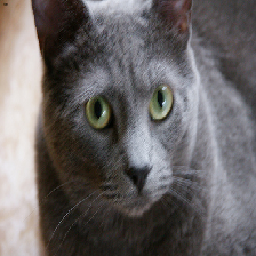} &
     \includegraphics[width=2cm]{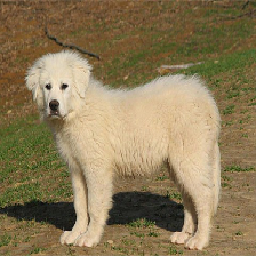} &
     \includegraphics[width=2cm]{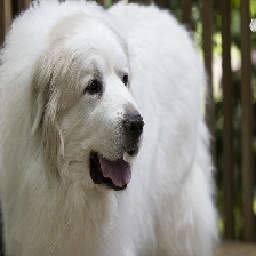} \\
     
     \toprule
     \includegraphics[width=2cm]{images/distances/test_image.png} &    
     \includegraphics[width=2cm]{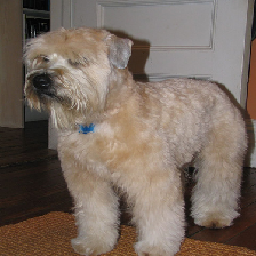} &
     \includegraphics[width=2cm]{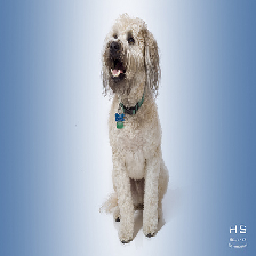} &
     \includegraphics[width=2cm]{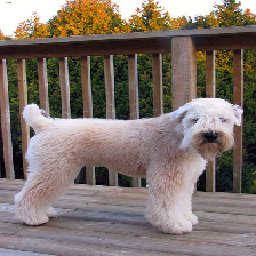} &
     \includegraphics[width=2cm]{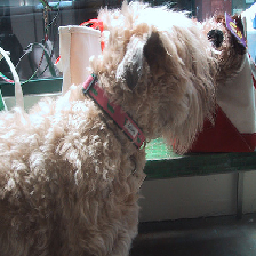} &
     \includegraphics[width=2cm]{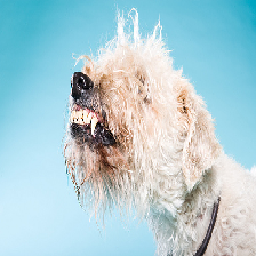} &
     \includegraphics[width=2cm]{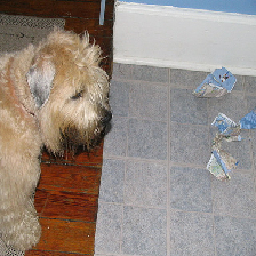} &
     \includegraphics[width=2cm]{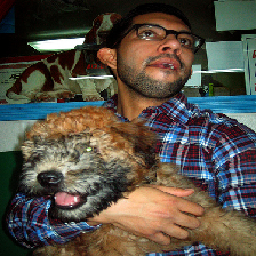} \\
     Cosine distance &    
     \includegraphics[width=2cm]{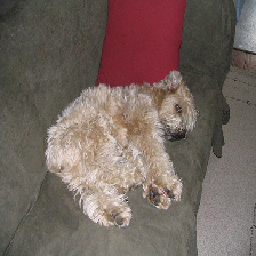} &
     \includegraphics[width=2cm]{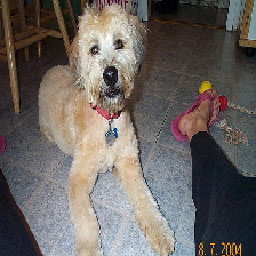} &
     \includegraphics[width=2cm]{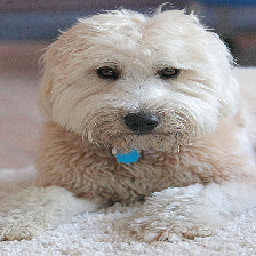} &
     \includegraphics[width=2cm]{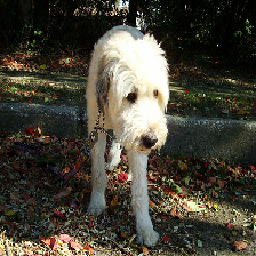} &
     \includegraphics[width=2cm]{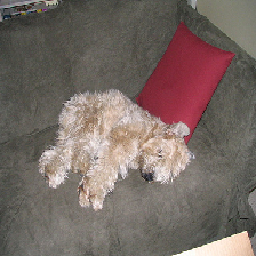} &
     \includegraphics[width=2cm]{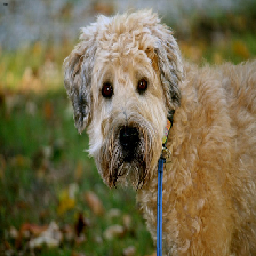} &
     \includegraphics[width=2cm]{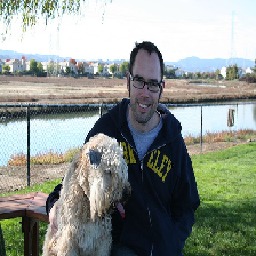} \\
     
     \toprule     
     \end{tabular}}
     \caption{\hl{Comparing the top $14$ images selected using different distance metrics in the embedding space for a given test image.}}
     \label{fig:similar_distances}
\end{table*}

\begin{table*}[!ht]
     \begin{center}
     \begin{tabular}{c c c c c}
     \toprule
     \includegraphics[width=0.175\linewidth]{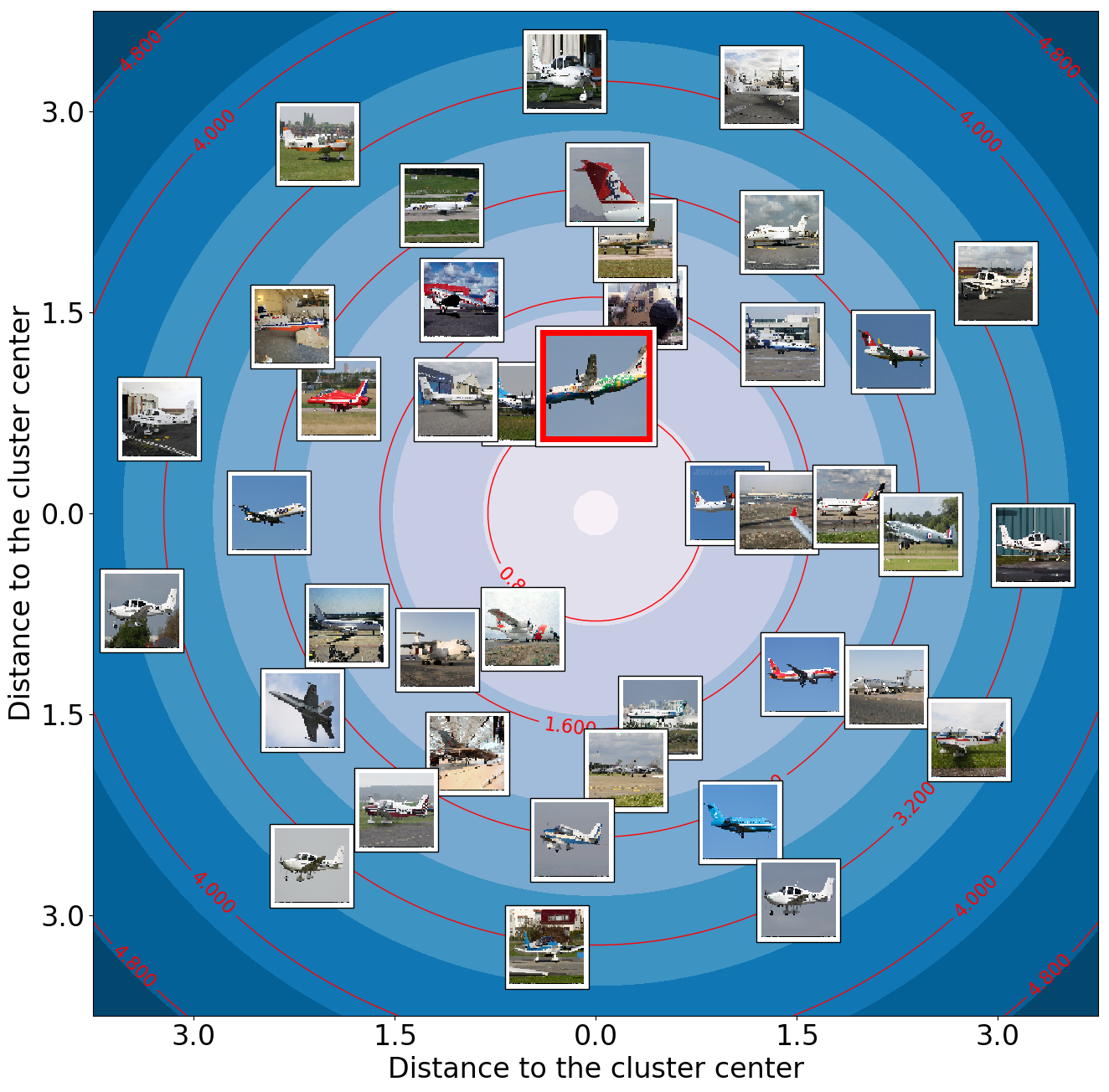} & \includegraphics[width=0.175\linewidth]{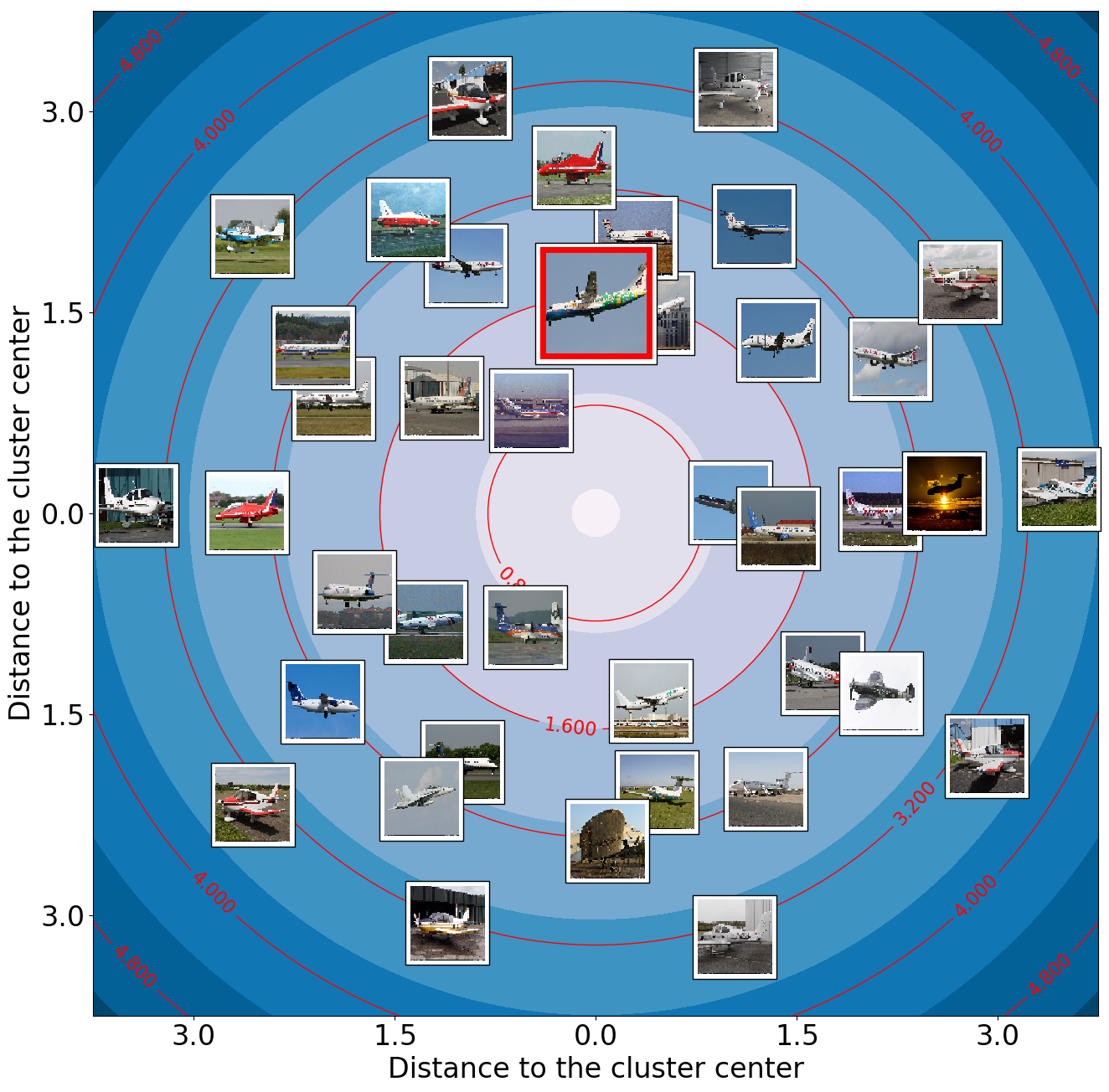} & \includegraphics[width=0.175\linewidth]{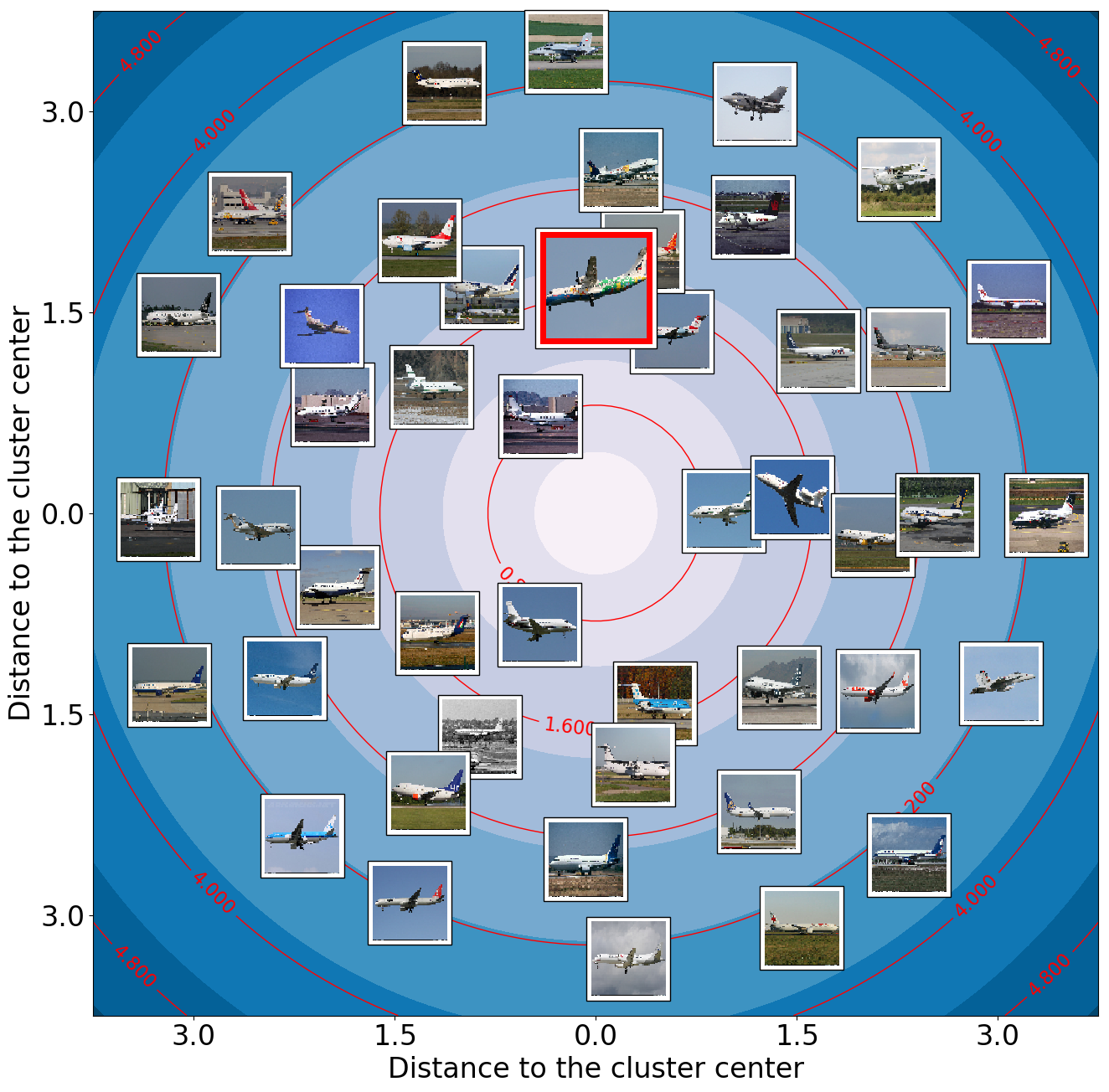} & \includegraphics[width=0.175\linewidth]{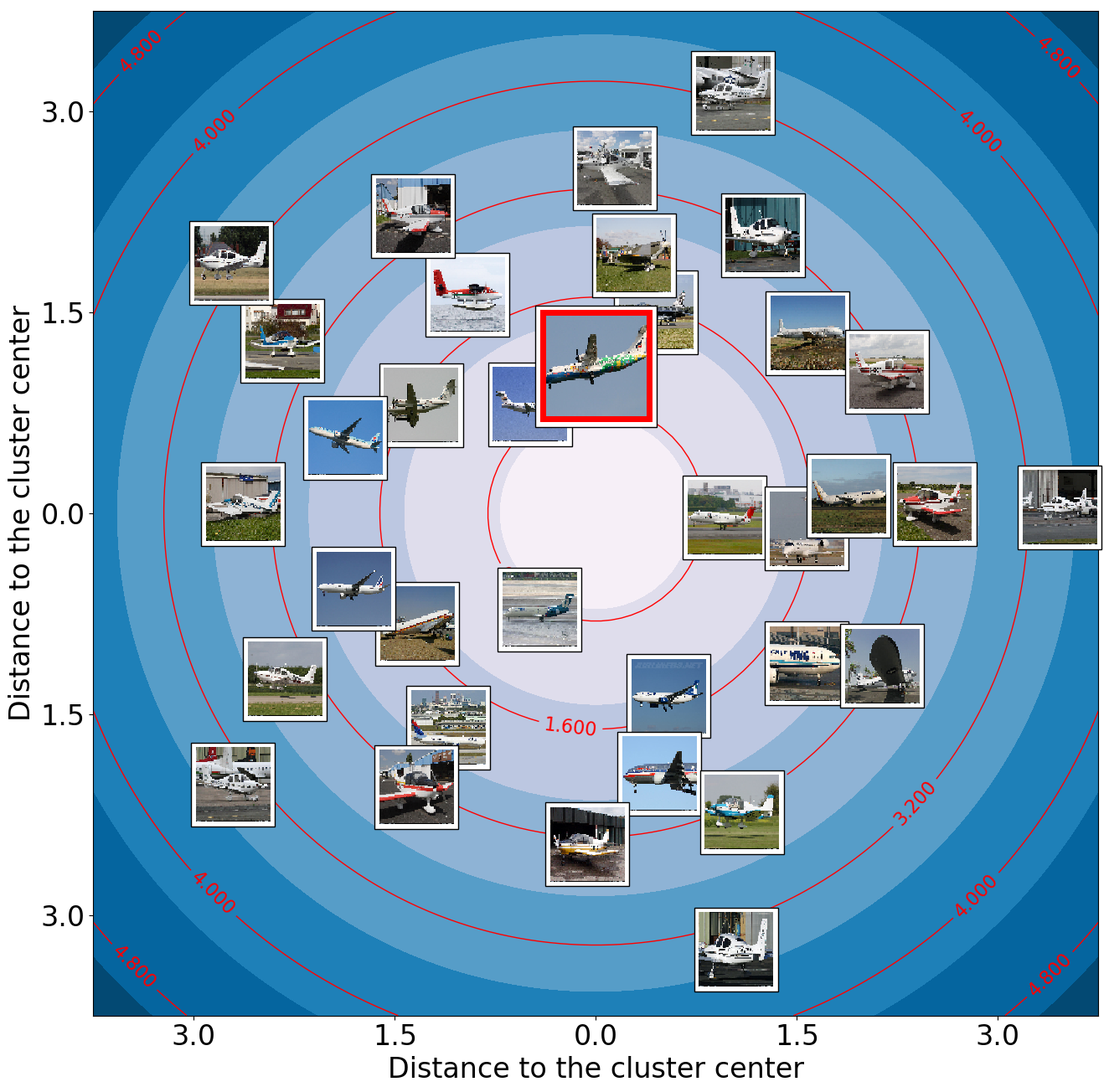} & \includegraphics[width=0.175\linewidth]{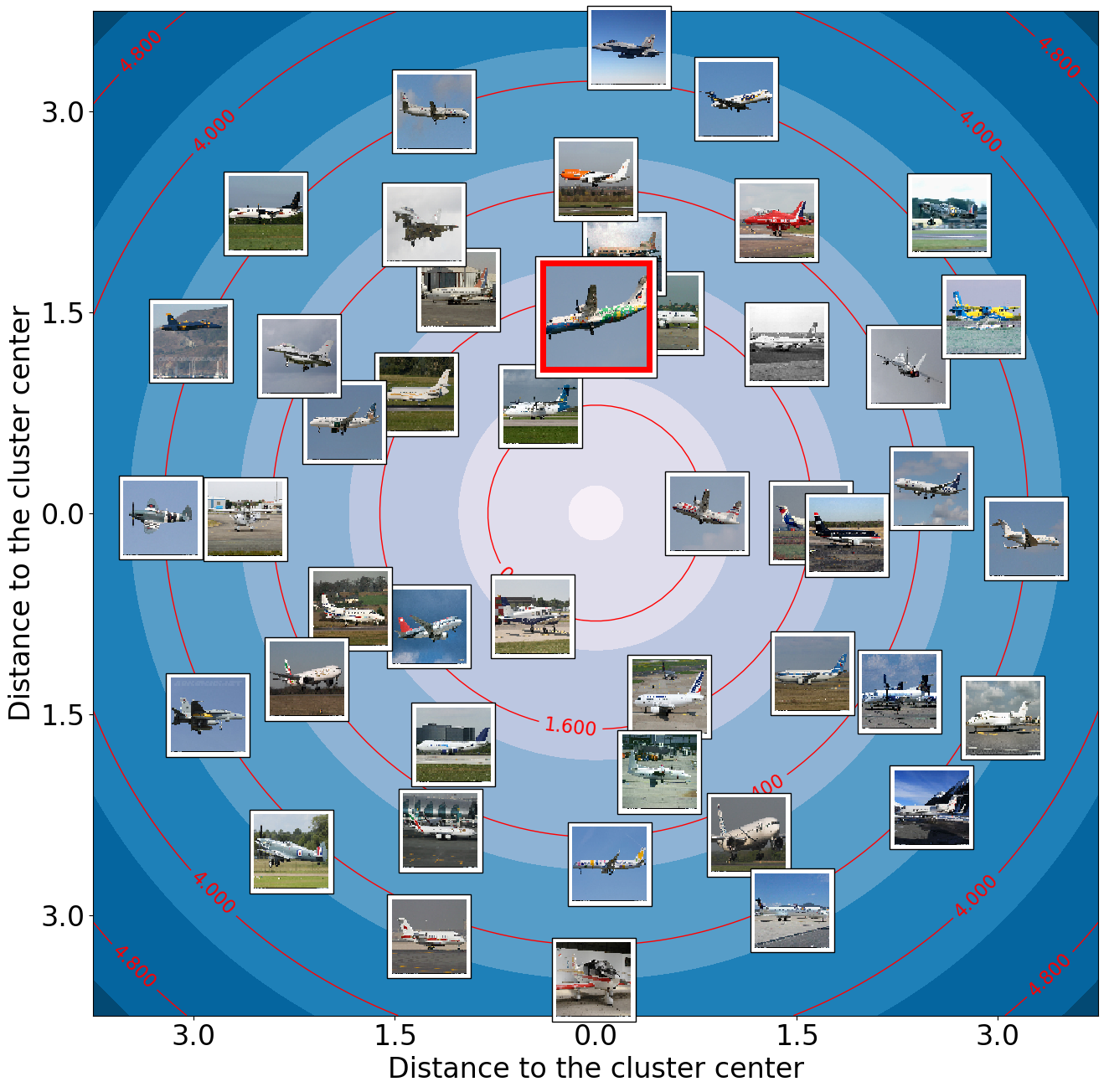} \\ \midrule
     \includegraphics[width=0.175\linewidth]{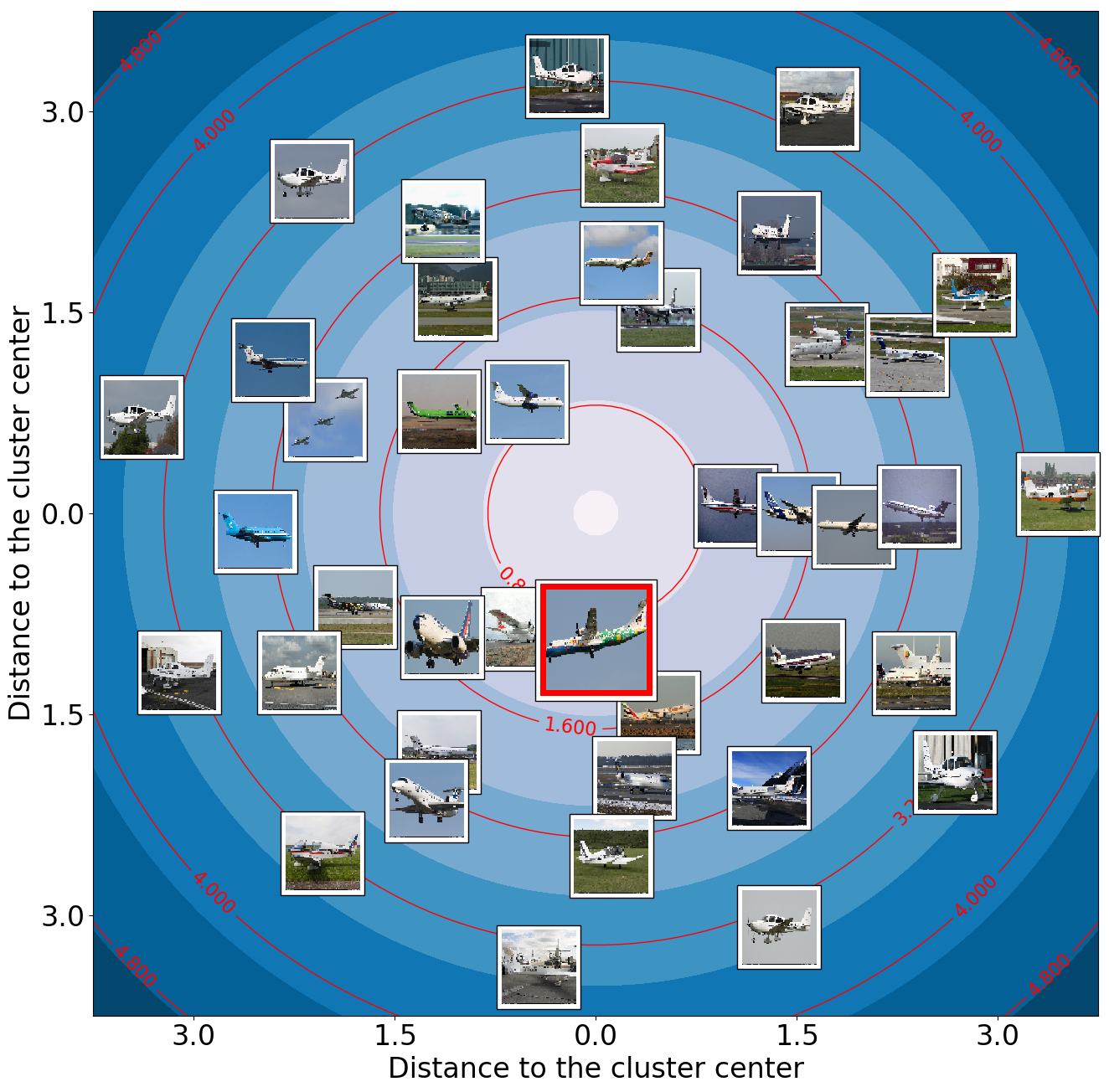} & \includegraphics[width=0.175\linewidth]{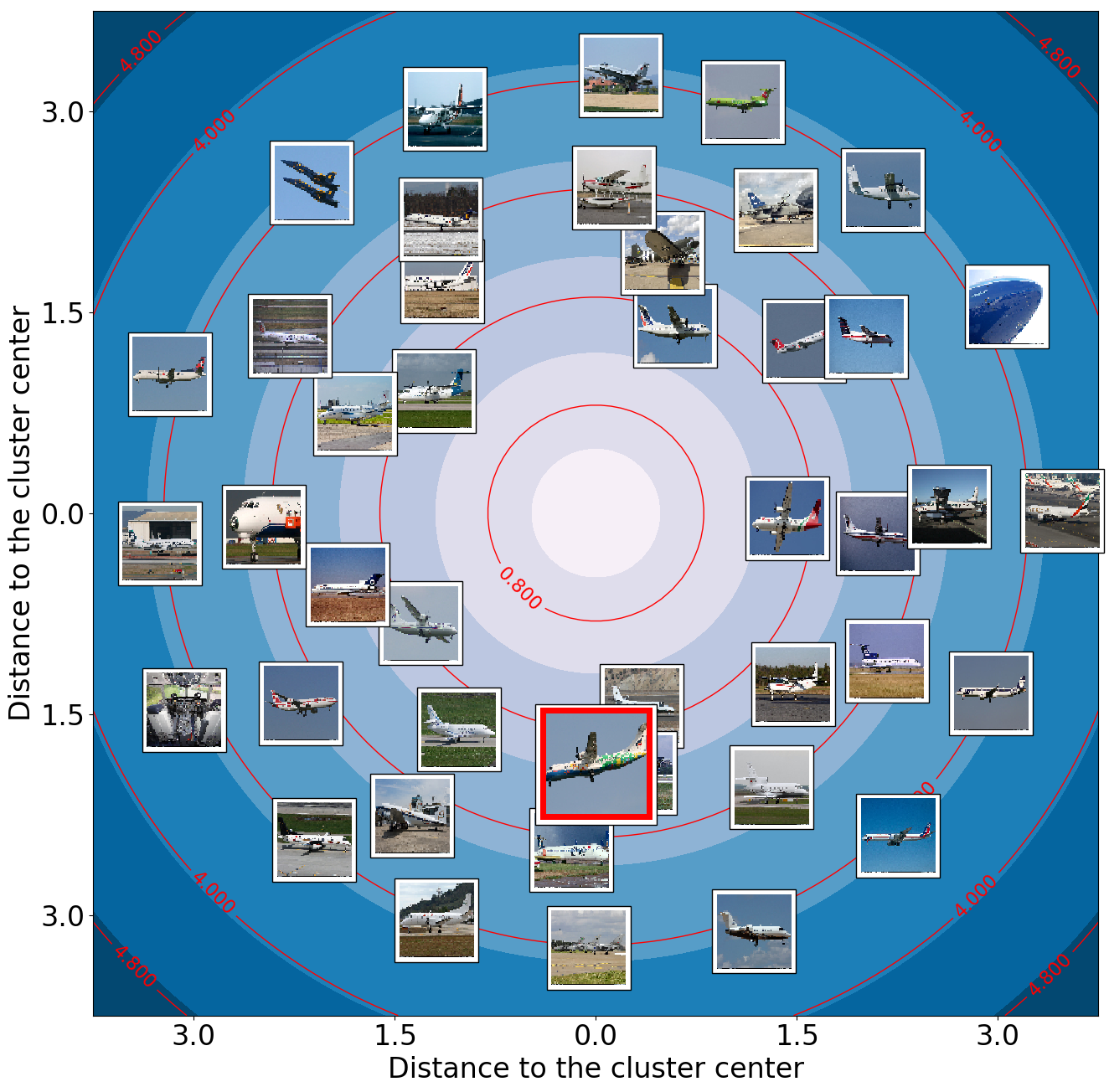} & \includegraphics[width=0.175\linewidth]{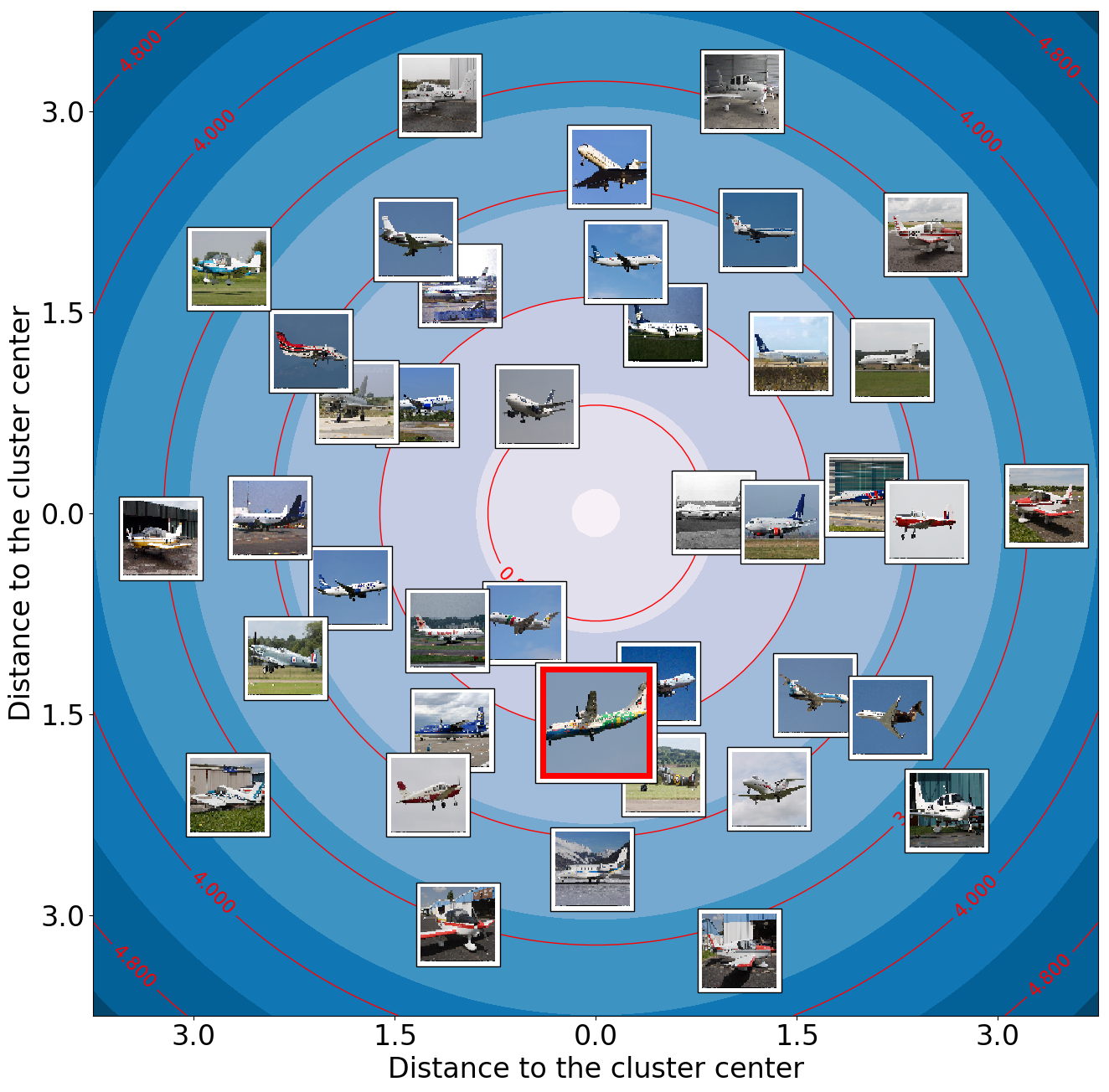} & \includegraphics[width=0.175\linewidth]{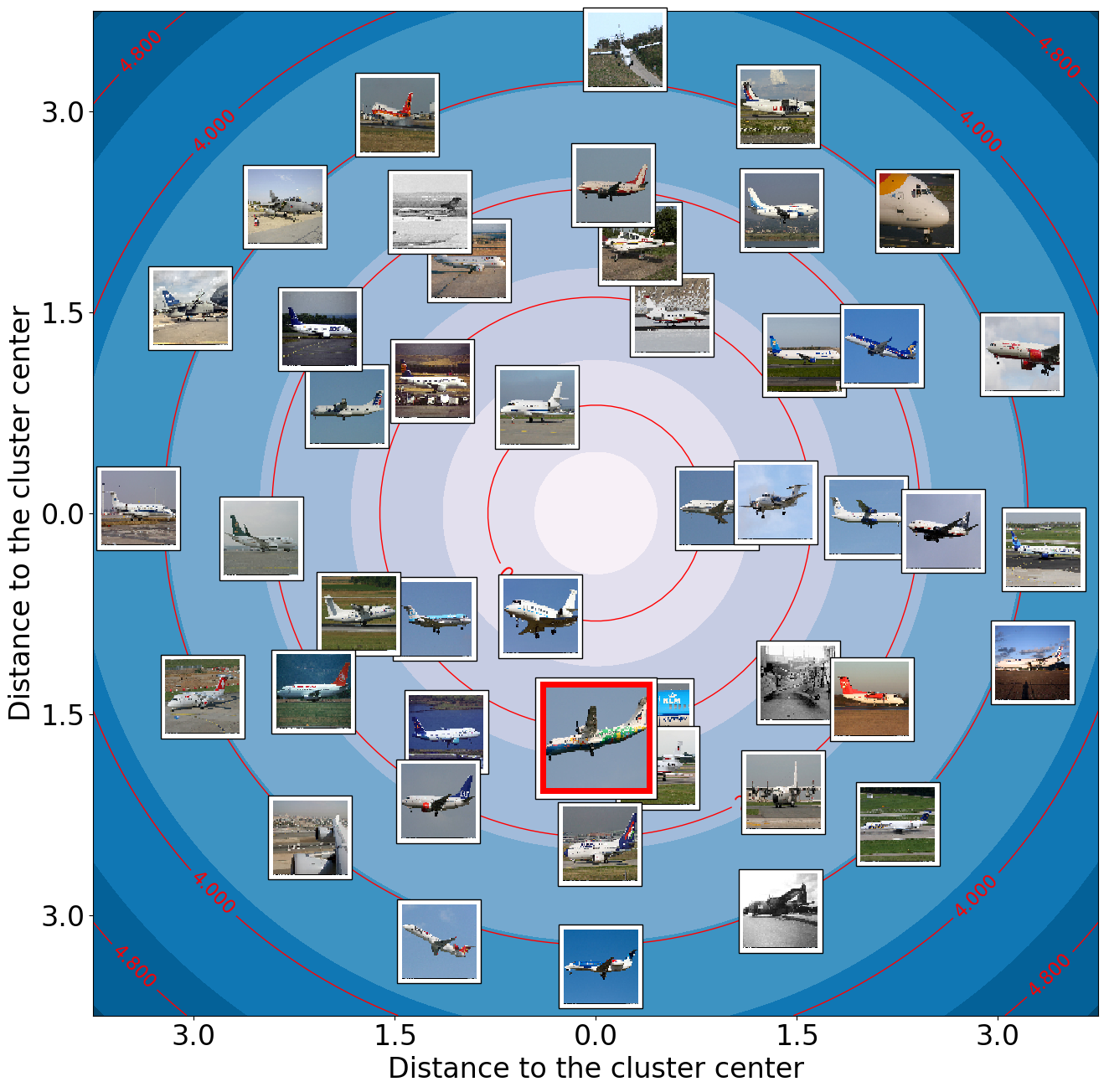} & \includegraphics[width=0.175\linewidth]{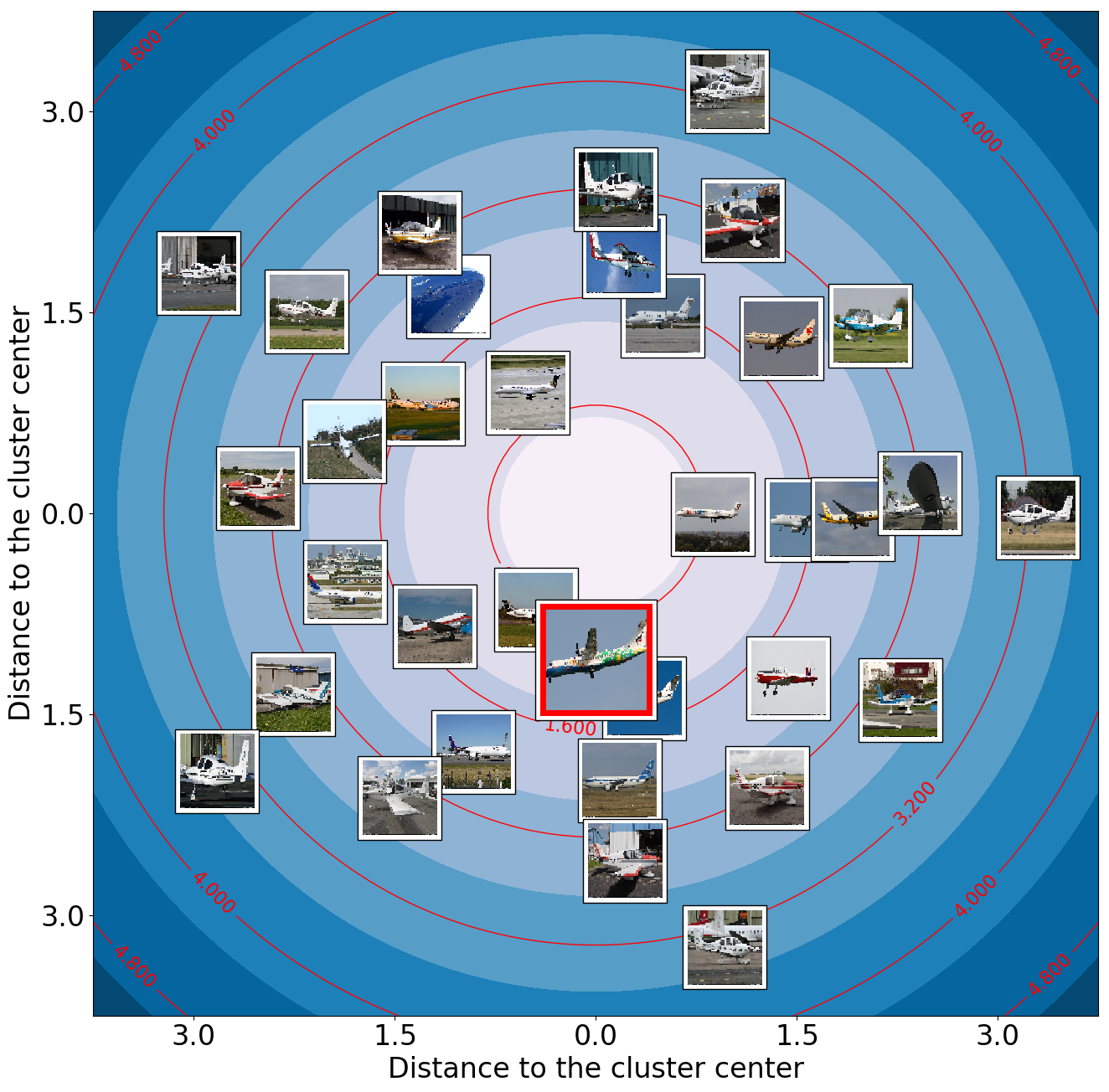} \\   
     \toprule
     \end{tabular}
     \end{center}
     \caption{The Figure illustrates the clusters contributing to the correct class (top row) and the wrong class (bottom row) of the CNN-RBF network. The test sample is the larger image with red boarded in each cluster representation. Red circles show the distance of the samples to the cluster center, and the background is proportional to the activation values of the cluster. The brighter the activation value, the larger it is and the maximum activation at the cluster center is equal to one.}
     \label{fig:corret_wrong_cluster}
\end{table*}

Using a different approximation strategy compared with fully connected layers provides CNN-RBFs with the chance to probe the decision-making process based on these visual clues:
\begin{itemize}
    \item Similar images as measured by the similarity distance metric of RBFs trained on CNN embeddings
    \item Visualizing the clusters with higher contribution to the network's decision and distance of the samples from the centers of these clusters
\end{itemize}

The embeddings of CNN are evaluated by their learned distance metric from cluster centers in RBFs. The same distance can be used to measure the distance between a test image and similar images from training data. Figure \ref{fig:similar_dissimilar} shows the similar images found in the training dataset for a given test sample by the similarity distance metric in Equation \ref{eq:distance}. The most similar and dissimilar images are computed using the following criteria:

\begin{flalign}\label{eq:similar}
\boldsymbol{x}_{similar} = \argmin_{\boldsymbol{x}_{train}} \parallel \boldsymbol{x}^\mu_{train}-\boldsymbol{x}_{test} \parallel^2_{\boldsymbol{R}}  &&
\end{flalign}
\begin{flalign}\label{eq:dissimliar}
\boldsymbol{x}_{dissimilar} = \argmax_{\boldsymbol{x}_{train}} \parallel \boldsymbol{x}^\mu_{train}-\boldsymbol{x}_{test} \parallel^2_{\boldsymbol{R}} &&
\end{flalign} 

where $x_{test}$ presents the input of RBFs for a given test image, $x^\mu_{train}$ shows the input vector for training samples, and $\mu$ enumerates the training samples from $1$ to $N$. $\boldsymbol{x}_{simliar}$ and $\boldsymbol{x}_{dissimliar}$ represent the most similar and dissimilar images to the given test image ($x_{test}$) respectively. We can use the same similarity metrics in Equation \ref{eq:similar} and \ref{eq:dissimliar} to create a ranked list of similar and dissimilar images for a given test sample.


\hl{Figure} \ref{fig:similar_distances} \hl{compares the performance of the similar sample selection for a given test images. The figure suggests that the learned metric and Euclidean distances outperform the cosine distance for similar sample selection. Furthermore the learned metric slightly outperforms the Euclidean distance in this specific case.}
 
The active clusters for every sample provide the reasoning behind the final decision of a CNN-RBF. The clusters can be described by the distance of images from their centers. Figure \ref{fig:corret_wrong_cluster} depicts training samples and their distances from the cluster centers \hl{against} a test sample. The \hl{product} of activations and output weights \hl{determines} the final decision of an RBF. Thus, the importance of clusters for a decision can be determined by sorting the \hl{product} of activations and class weights. Figure \ref{fig:corret_wrong_cluster} depicts the clusters with the highest contributions to the correct class (ground truth) and the wrong class based on \hl{this multiplication product}. The wrong class here refers to the class with the second-highest level of confidence.
\section{Conclusion}
\label{sec:conclusion}
The research work presents fundamental architectural modifications that are applied to \hl{RBFs to integrate them with CNNs} for computer vision. The experimental results indicate that the integration of RBFs on top of CNNs achieves competitive performances in benchmark computer vision datasets by combining supervised and unsupervised learning. The proposed activation and training process is compatible with any arbitrary state-of-the-art CNN architecture, including inception blocks and residual connections. The small gap between the CNN-RBFs performance and best CNN models is a subject \hl{for} future research to find optimal regularization methods for RBF networks. Using RBF architecture with CNNs introduces two unique and network-specific opportunities for learning a similarity distance metric and interpreting the decision-making process in more detail. Similar and dissimilar images found using a similarity distance metric trained by RBFs are interpretable by humans. \hl{The cluster representations are currently only used to trace the decision making process as in the current research, the distribution of images around clusters are not visually conclusive due to being optimized in an unsupervised manner regardless of ground-truth labels.}

\section*{\hl{Acknowledgments}}
\hl{We acknowledge Thilo Stadelmann and Yvan Putra Satyawan for their suggestions, grammatical and stylistic edits.
This paper was made possible by the open source Tensorflow}~\cite{tensorflow2015-whitepaper} \hl{and Keras}~\cite{chollet2015keras} \hl{deep learning libraries, the Weights \& Biases toolbox}~\cite{wandb}\footnote{\href{https://www.wandb.com/}{\url{https://www.wandb.com/}}} \hl{for experiment logging, and PlotNeuralNet}\footnote{\href{https://github.com/HarisIqbal88/PlotNeuralNet}{\url{https://github.com/HarisIqbal88/PlotNeuralNet}}} \hl{for neural network visualization. We acknowledge the significant effort invested into developing these tools.}

\bibliography{main}
\bibliographystyle{IEEEtran}

\begin{IEEEbiography}[{\includegraphics[width=1in,height=1.25in,clip,keepaspectratio]{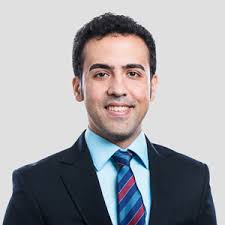}}]{\textbf{M\lowercase{ohammadreza} A\lowercase{mirian}}} received a M.Sc. \mbox{degree} in Communications Technology from Ulm University, Ulm, Germany, in 2017. He is \mbox{currently} working as a researcher at the \mbox{Institute} of Applied Information Technology (InIT) of the Zurich University of Applied Sciences (ZHAW), Winterthur, Switzerland and \mbox{simultaneously} \mbox{pursuing} a Ph.D. degree at Ulm University. \mbox{Besides} his research interests in biophysiological signal processing for person-centered medical and \mbox{affective} pattern recognition, his current research focuses on interpretable deep-learning algorithms for industrial applications and automated deep learning. 
\end{IEEEbiography}
\newpage
\begin{IEEEbiography}[{\includegraphics[width=1in,height=1.25in,clip,keepaspectratio]{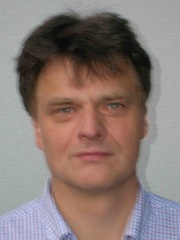}}]{\textbf{F\lowercase{riedhelm} S\lowercase{chwenker}}} studied mathematics and computer science at the University of \mbox{Osnabr{\"u}ck}, where he received his diploma and Ph.D. degree. Currently, he is a Privatdozent at Ulm University, Institute of Neural Information Processing. His research interests include - but are not limited to - artificial neural networks, machine learning, statistical learning theory, data mining, pattern recognition, information fusion and \mbox{affective} computing. He (co-)edited 20 special issues and workshop proceedings published in international journals and publishing companies. He published 200+ papers at international conferences and journals and served as \mbox{(Co-)Chair} of the IAPR TC3 on {Neural Networks and Computational Intelligence}. Since 2016 he is the Chair of the new IAPR TC9 on {Pattern Recognition in Human-Computer Interaction}.
\end{IEEEbiography}


\EOD

\end{document}